\documentclass[numbib]{cta-author}

{}
{}
{}

\usepackage{tikz}
\usetikzlibrary{arrows}
\usetikzlibrary{shapes,positioning}
\usetikzlibrary{tikzmark}
\tikzset{>=latex}
\usepackage[export]{adjustbox}
\graphicspath{{img/}}
\usepackage{nicematrix}
\usepackage{booktabs}
\usepackage{amsmath,amssymb,amsthm}	
\usepackage{mathtools}	
\usepackage{colortbl}
\usepackage{bm} 
\usepackage{algorithm}
\usepackage{algpseudocode}  
\usepackage{algorithmicx}
\usepackage{comment}
\usepackage{caption}
\usepackage{subcaption}
\usepackage{hyperref}
\usepackage{ifsym}
\usepackage[switch,columnwise]{lineno}
\usepackage{fancyhdr}

\AtBeginDocument{
	\label{CorrectFirstPageLabel}
	
}

\definecolor{aliceblue}{rgb}{0.94, 0.97, 1.0}
\clearemptydoublepage

\begin{document}

\title{Worst-Case Morphs using Wasserstein ALI and Improved MIPGAN}

\author{\au{U.M. Kelly$^{1\correnvelope}$, }
\au{M. Nauta$^1$, }
\au{L. Liu$^1$, }
\au{L.J. Spreeuwers$^1$, }
\au{R.N.J. Veldhuis$^{1,2}$}
}

\address{\add{1}{Data Management and Biometrics Group, Faculty of EEMCS, University of Twente, 7500 AE Enschede, The Netherlands}
\add{2}{Department of Information Security and Communication Technology, Norwegian University of Science and Technology, Gjøvik, Norway}
\email{ u.m.kelly@utwente.nl}} 

\begin{abstract}
A morph is a combination of two separate facial images and contains identity information of two different people. When used in an identity document, both people can be authenticated by a biometric Face Recognition (FR) system. Morphs can be generated using either a landmark-based approach or approaches based on deep learning such as Generative Adversarial Networks (GAN). In a recent paper, we introduced a \textit{worst-case} upper bound on how challenging morphing attacks can be for an FR system. The closer morphs are to this upper bound, the bigger the challenge they pose to FR. We introduced an approach with which it was possible to generate morphs that approximate this upper bound for a known FR system (white box), but not for unknown (black box) FR systems. 

In this paper, we introduce a morph generation method that can approximate worst-case morphs even when the FR system is not known. A key contribution is that we include the goal of generating difficult morphs \emph{during} training. Our method is based on Adversarially Learned Inference (ALI) and uses concepts from Wasserstein GANs trained with Gradient Penalty, which were introduced to stabilise the training of GANs. We include these concepts to achieve similar improvement in training stability and call the resulting method Wasserstein ALI (WALI). We finetune WALI using loss functions designed specifically to improve the ability to manipulate identity information in facial images and show how it can generate morphs that are more challenging for FR systems than landmark- or GAN-based morphs. We also show how our findings can be used to improve MIPGAN, an existing StyleGAN-based morph generator. 

\end{abstract}

\maketitle

\section{Introduction}
It has been shown that \textit{morphing attacks} pose a significant risk to both Face Recognition (FR) systems and humans, e.g. border guards \cite{SNR17, Robertson2017}. A morph is an image that is created by combining two images of two different people. If it contains sufficient identity information of each person, then FR systems and humans will accept the morph as a match both when it is compared to a different image of the first person, but also when it is compared with a different image of the second person. This means that two people could share one passport or other identity document and avoid travel restrictions or border controls, e.g. a criminal could travel using the identity document of an accomplice. Some countries intend to stop allowing people to bring their own printed photos for passport applications, e.g. Germany \cite{linkDE}. At the same time, there are still countries that allow applicants to provide their own digital or printed passport photo, e.g. Ireland \cite{linkEIR}. Morphed images also pose a challenge in other scenarios since two people could for example share a driver's licence, health insurance, public transportation tickets etc. There are myriad ways to exploit systems, subscriptions, access rights, and more using morphed images.

Generative Adversarial Networks (GAN) have been shown to successfully generate fake data that matches a real data distribution \cite{GAN}. Image characteristics such as expression or age (in the case of facial images) can be manipulated by applying changes to latent representations of images, which are vectors in a GAN's latent space. If an inversion were available that maps images to the latent space of a GAN, then this would allow advantage to be taken of the benefits that GANs provide and allow real data to be manipulated directly. Mapping two images onto two respective latent vectors, and finding an appropriate interpolation between those two vectors would then lead to a GAN-generated morph. Both MorGAN \cite{MorGAN} and MIPGAN \cite{MIPGAN} are examples of this approach.

Morph generation can rely on landmark-based, GAN-based or manual methods. More recently, morphs generated using diffusion models were introduced \cite{MorDiff, blasingame2023leveraging}. How challenging morphs are varies depending on implementation details such as the landmark detector used, the splicing method used, post-processing, whether images were printed and scanned, which pairs of images were selected for morphing, etc. A criminal could make a morph using hand-selected landmarks, and then iteratively apply changes and test the morph using one or more FR systems to find a morph that is most likely to be accepted by FR systems. They could also apply changes that make it harder for Morphing Attack Detection (MAD) methods to detect the morphs. This means that the variation in morphing methods used in research may not be representative of morphs that could exist in reality, since criminals will not advertise which morphing methods they are using. Therefore, the estimated vulnerability of FR and MAD systems may be different on such morphs than on datasets generated by researchers, where some trade-off between quantity and quality may have to be made.

MAD methods have been proposed, targeted at detecting landmark-based morphs, GAN-based morphs or both. Developing an MAD approach that can detect both landmark- and GAN-based morphs - especially if they are of a type not seen during training - is still an open challenge \cite{RAJA2022104535}. GAN-based morphing detection is very similar to the general detection of GAN images (deepfakes) \cite{ColboisGAN}. Increasing the variation in available morphing tools could be helpful in the development of detection methods, since both in GAN-based morph detection and deepfake detection more generally, it has been shown that methods struggle to detect images of a type not seen during the training phase.

In \cite{wcMorphing} it was shown that theoretically - and if the FR system is known also in practice - morphs can be even more challenging than either landmark- or GAN-morphs. While landmark-based morphing combines images in the image domain, GAN-based morphing combines them by mapping them to embeddings in the GAN latent space, interpolating in that latent space, and generating a morph from the interpolated latent embedding. On the other hand, our approach in \cite{wcMorphing} was to directly reverse the mapping from images to latent embeddings in the FR latent space (different from the GAN latent space). This approach can be used to exploit the vulnerabilities of the FR system it was trained with, but is less suited than GAN-based methods to generate morphs that visually (to humans) look like both contributing identities and struggles to fool unseen FR systems.

In this work, we continue this investigation to find out whether it is possible to automatically generate morphs that approximate the theoretical \textit{worst case} for more than one face recognition system simultaneously, even when the FR system is unknown (``black box''), showing there are morphs that can be even more challenging than landmark- or GAN-based morphs. The variation of morphs used in existing MAD benchmarks, such as \cite{raja2020morphing_short,NIST,BOEP}, can be increased by including approximations of worst-case morphs.\\

Our contributions consist firstly of adapting the method introduced in Adversarially Learned Inference (ALI) \cite{dumoulin2017adversarially} and improving it to better enable manipulation of real data, e.g. generating interpolations of real images. We call the resulting improved method Wasserstein ALI (WALI) and use it to generate morphs. Like ALI, WALI jointly learns a generative and an inverse mapping, enabling its use for morph generation. We improve training stability, which allows generation of larger images: we generate images using WALI of up to 512$\times$512 pixels, compared to 64$\times$64 pixels achieved by ALI. It may be possible to generate images with even higher resolutions using WALI, but we did not try this, due to hardware and time restraints. ALI's aim is to generate images that look as real as possible, which means it is not necessarily optimal for generating \textit{challenging} morphs. WALI is further improved for this purpose by including loss functions designed specifically to improve the ability to manipulate identity information in facial images. The resulting model provides an easy way to generate (large) morphing datasets intended for training or evaluating face recognition and morphing attack detection systems.

Our second set of contributions lies in applying WALI and our improved implementation of MIPGAN to approximate worst-case morphs, evaluating these approximations, and comparing them to other morphs. Since morphs generated using an underlying StyleGAN Generator \cite{StyleGAN_FlickR} are currently the SOTA when it comes to GAN-based morphing, we include MIPGAN morphs in all our comparisons. 
Summarising, our main contributions are
\begin{itemize}
	\item Improving ALI to enable morph generation, resulting in Wasserstein ALI (WALI), which provides an easy way to generate (large) morphing datasets intended for training or evaluating face recognition and morphing attack detection systems,
	\item showing that already considering the goal of generating difficult morphs \textit{during} training instead of only during optimisation (after training) leads to more challenging morphs in both white-box and black-box settings than if WALI is only trained to generate real-looking images,
	\item showing that optimisation on our trained model leads to morphs that are more challenging for FR systems than landmark- or MIPGAN-morphs, even when evaluating under black-box settings. This proves the existence of morphs that lie closer (than landmark or MIPGAN) to the theoretical worst-case morph for six out of eight FR systems we evaluated,
	\item showing that optimising towards a worst-case embedding is also possible when using existing generative models. Since we see that WALI does not generalise well to new datasets that are different from the data it was trained on, we also apply some of our suggested improvements to a StyleGAN Generator that is better at generalising to new datasets, resulting in an improved MIPGAN approach that also leads to more challenging morphs than other GAN-based approaches.
\end{itemize}

\section{Related Work} \label{RelatedWork}

\subsection{Worst-Case Morphs}
In \cite{wcMorphing} an upper bound on the vulnerability of FR systems to morphing attacks was introduced. Let $\varphi$ be the function that describes an FR system's mapping from the image space $X$ to the embedding space $Y$, i.e. $\varphi: X \rightarrow Y$. If $d$ is the dissimilarity score function that is used to calculate the dissimilarity score for pairs of embeddings in $Y$, then the \textit{worst-case embedding} for two images $\bm{x}_1$ and $\bm{x}_2$ is
\begin{equation}
\bm{y}^* := \text{argmin}_{\bm{y} \in Y} \left( \max \left[ d(\bm{y},\varphi(\bm{x}_1)), d(\bm{y},\varphi(\bm{x}_2)) \right] \right). \label{eq:z_wc} 
\end{equation}

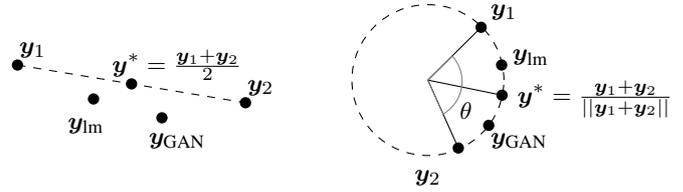
\begin{figure}[h]
	\centering
	\begin{tikzpicture}
	\draw[dashed, black, -] (0,0) -- (3,-0.5);
	
	\node at (0,0) [circle,fill,inner sep=1.5pt]{};
	\node[] at (0.2,0.2)
	{\text{$\bm{y}_1$}};
	
	\node at (3,-0.5) [circle,fill,inner sep=1.5pt]{};
	\node[inner sep=0pt] (whitehead) at (3.2,-0.3)
	{\text{$\bm{y}_2$}};
	
	\node at (1.5,-0.25) [circle,fill,inner sep=1.5pt]{};
	\node[inner sep=0pt] (whitehead) at (2.1,-0.0)
	{\text{$\bm{y}^* = \frac{\bm{y}_1+\bm{y}_2}{2}$}};
	
	\node at (1.0,-0.45) [circle,fill,inner sep=1.5pt]{};
	\node[inner sep=0pt] (whitehead) at (0.9,-0.8)
	{\text{$\bm{y}_{\text{lm}}$}};
	
	\node at (1.9,-0.7) [circle,fill,inner sep=1.5pt]{};
	\node[inner sep=0pt] (whitehead) at (2.1,-1.0)
	{\text{$\bm{y}_{\text{GAN}}$}};
	
	\draw[black,dashed] (5.4,-0.2) circle (1cm);
	
	\node at (6.1,0.5) [circle,fill,inner sep=1.5pt]{};
	\node[] at (6.4,0.7)
	{\text{$\bm{y}_1$}};
	
	\node at (5.8,-1.1) [circle,fill,inner sep=1.5pt]{};
	\node[] at (5.4,-1.5)
	{\text{$\bm{y}_2$}};
	
	\node at (6.375,-0.4) [circle,fill,inner sep=1.5pt]{};
	\node[] at (7.6,-0.45)
	{\text{$\bm{y}^* = \frac{\bm{y}_1+\bm{y}_2}{||\bm{y}_1+\bm{y}_2||}$}};
	
	\draw[black, -] (5.4,-0.2) -- (6.1,0.5);
	\draw[black, -] (5.4,-0.2) -- (5.8,-1.1);
	\draw[black, -] (5.4,-0.2) -- (6.375,-0.4);
	
	\node[] at (5.9,-0.6)
	{\text{$\theta$}};
	\draw[gray] (5.4,-0.2) -- (5.6,-0.65) arc [start angle=-60, delta angle=102, radius=0.5cm] -- (5.4,-0.2);
	
	\node at (6.37,-0.0) [circle,fill,inner sep=1.5pt]{};
	\node[inner sep=0pt] (whitehead) at (6.75,0.1)
	{\text{$\bm{y}_{\text{lm}}$}};
	
	\node at (6.2,-0.8) [circle,fill,inner sep=1.5pt]{};
	\node[inner sep=0pt] (whitehead) at (6.6,-1.0)
	{\text{$\bm{y}_{\text{GAN}}$}};
	
	\end{tikzpicture}
	\caption{\label{worstcase}The worst-case embedding $\bm{y}^*$ when $d$ denotes euclidean distance (left) or angle (right). If it exists, an image that maps to $\bm{y}^*$ is even more challenging than a landmark- ($\bm{y}_{\text{lm}}$) or GAN-based morph ($\bm{y}_{\text{GAN}}$).}
\end{figure}

For example, if $d$ returns the euclidean distance, denoted as $||.||_2$, between two embeddings $\bm{y}_1$ and $\bm{y}_2$, then the dissimilarity score is $d(\bm{y}_1, \bm{y}_2) = ||\bm{y}_1-\bm{y}_2||_2$. In that case $\bm{y}^*$ is that $\bm{y}$ for which $d(\bm{y}_1,\bm{y})=d(\bm{y},\bm{y}_2)=d(\bm{y}_1, \bm{y}_2)/2$, see the example on the left in Fig.~\ref{worstcase}.

If an FR system uses similarity scores, defined by a function $S$, then
\begin{equation}
	\bm{y}^* := \text{argmax}_{\bm{y} \in Y} \left( \min \left[ S(\bm{y},\varphi(\bm{x}_1)), S(\bm{y},\varphi(\bm{x}_2)) \right] \right). \label{z_wc_sim } 
\end{equation}
For example, if $S$ returns cosine similarity, then $S(\bm{y}_1, \bm{y}_2) =\cos(\theta)$, where $\theta$ is the angle between $\bm{y}_1$ and $\bm{y}_2$, see Fig.~\ref{worstcase}. In that case $\bm{y}^*$ is any $\bm{y}$ for which $S(\bm{y}_1,\bm{y})=S(\bm{y},\bm{y}_2)=\cos(\theta/2)$.

Since worst-case embeddings can be calculated using only normal (bona fide) images, no morphs are needed to compute the worst-case upper bound. This means that the potential vulnerability of an FR system can be determined without having to make or evaluate one single morph.

\subsection{GANs for Morph Generation}

MorGAN \cite{MorGAN} uses Adversarially Learned Inference (ALI) to generate $64\times64$ pixel morphs. ALI consists of training three networks: an Encoder, a Decoder (similar to the Generator in a plain GAN) and a Discriminator. MorGAN generates morphs by passing two images through the Encoder, interpolating between the two resulting latent embeddings and then passing this interpolation through the Decoder. This approach results in an image that shares similarities with both original images. Resulting morphs have low resolution and compared to landmark-based morphs are not nearly as successful at fooling FR systems.

MIPGAN \cite{MIPGAN} makes use of a pretrained StyleGAN network by training an Encoder that encodes images into the StyleGAN latent space. Optimisation is then used to approximate an optimal embedding in the StyleGAN latent space, that when passed through StyleGAN results in a morph. The morphs are visually convincing, as confirmed by studies on human ability to distinguish between morphs and real images. They are about as successful at attacking FR systems as landmark-based morphs. 
The MIPGAN method is improved on in RegenMorph \cite{Regenmorph}. The resulting images are visually more convincing, but are shown to be less successful than MIPGAN morphs at fooling FR systems.

What these existing GAN-based images have in common, is that the underlying networks were all trained with the goal of generating fake images that look like real images. While MorGAN uses a pixel-based loss to preserve identity in images, none of the networks were specifically \textit{trained} to generate morphs. This means that optimisation may be used together with a trained and frozen network to find the optimal latent embedding that leads to a successful morph, but we hypothesise that already considering the goal of generating morphs \textit{during} instead of only \textit{after} training might lead to more successful morphs. Morphing attacks generated specifically to exploit vulnerabilities of deep-learning-based FR can be considered as a type of \emph{adversarial attack} on an FR system \cite{AdversarialAttacks}, since images are manipulated in a way similar to \emph{impersonation attacks}, where in the case of morphing, two identities are being ``impersonated'' simultaneously.

An overview of research on GAN inversion is provided in \cite{xia2021gan}, where new inverse networks are trained to invert already existing GANs. On the other hand, approaches such as in \cite{dumoulin2017adversarially,donahue2017adversarial} attempt to \textit{jointly} train an Encoder, a Decoder (the GAN Generator) and a Discriminator network. As mentioned in \cite{dumoulin2017adversarially}, it is possible that there are interactions that can be better learned by training these networks jointly, since the Encoder and Decoder can interact during training, which is not possible when using a frozen GAN. For this reason, we explore whether it is possible to improve methods that use the second approach, such as \cite{dumoulin2017adversarially,donahue2017adversarial}, by addressing some disadvantages such as unstable training. We show that the resulting approach Wasserstein ALI (WALI) is well-suited to approximate worst-case morphs.

\subsection{Morphing Attack Detection (MAD)}
Research on variation in morphing generation algorithms includes: post-processing landmark-based morphs to mask effects caused by the morphing process \cite{8897214}, a model to simulate the effects of printing and scanning \cite{FFM21} and considering the influence of ageing on morphing attacks \cite{9304856}.
The lack of variation in morphing techniques is addressed in \cite{SKR20}, which presents a method for MAD and evaluates it on morphs created using different algorithms, which are all landmark-based. Printed-and-scanned morphs are included in this evaluation, but GAN morphs or other methods are not taken into consideration.

In this work, we evaluate morphs using two MAD methods to show that if they are trained with landmark-based morphs only, then they struggle to detect WALI- as well as (improved) MIPGAN-based morphs, emphasising the need for varied datasets for training MAD.

\begin{figure*}
	\centering
	\includegraphics[width=0.8\paperwidth]{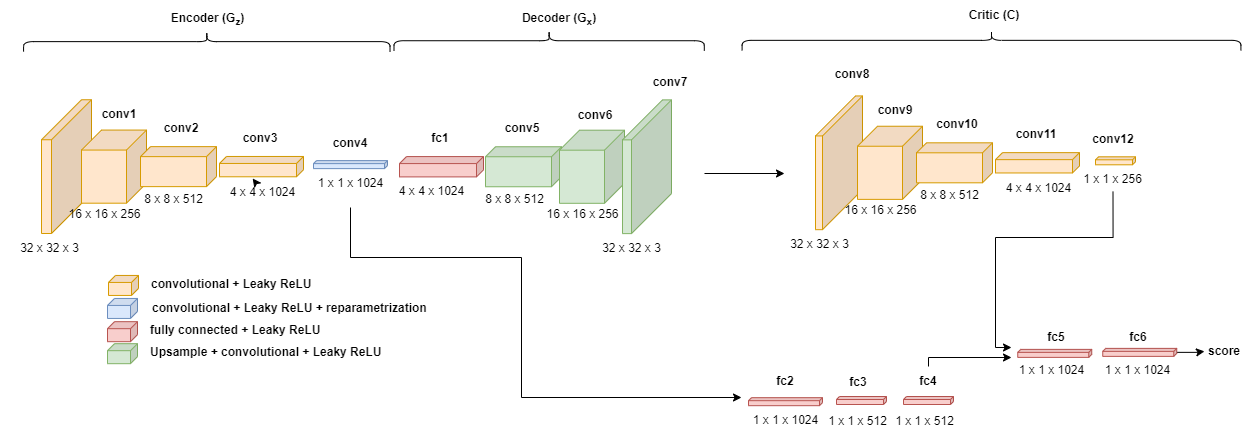}
	\caption{\label{WALI}The networks and architecture used in Wasserstein ALI (WALI).}
\end{figure*}

\section{Proposed System} \label{methodology}

\subsection{Adversarially Learned Inference}
In ALI \cite{dumoulin2017adversarially} two probability distributions over $\bm{x}$ and $\bm{z}$ are considered:
\begin{itemize}
	\item the Encoder joint distribution $q(\bm{x}, \bm{z}) = q(\bm{x})q(\bm{z} \mid \bm{x})$,
	\item the Decoder joint distribution $p(\bm{x}, \bm{z}) = p(\bm{z})p(\bm{x} \mid \bm{z})$.
\end{itemize}
The Encoder marginal $q(\bm{x})$ is the empirical data distribution over the image space $\mathcal{X}=[0,1]^{d_1}$, where $d_1 = w\times h \times n_c$, the width by height of the image by the number of colour channels $n_c$. The Decoder marginal $p(\bm{z})$ over the latent space $\mathcal{Z}$ is the distribution from which input noise is sampled, e.g. a standard Normal distribution $p(\bm{z}) = \mathcal{N} (0, I)$ over $\mathcal{Z}=(-\infty,\infty)^{d_2}$ (this can be truncated to $[-R,R]^{d_2}, \ R \in \mathbb{R}$ to ensure that $\mathcal{Z}$ is compact, which is needed to prove that ALI converges). Embeddings in the ALI latent space $\mathcal{Z}$ are denoted $\bm{z}$ and should not be confused with embeddings $\bm{y}$ in the FR latent space.

The objective of ALI is to match the two joint distributions. In order to achieve this, an adversarial game is played using:
\begin{itemize}
	\item $G_z$: an Encoder that maps from image space to a latent space,
	\item $G_x$: a Decoder that maps from the latent space to image space,
	\item $D$ (or $C$): \ a Discriminator (or Critic) that tries to determine whether joint pairs $(\bm{x}, \bm{z})$ are drawn either from $q(\bm{x}, \bm{z})$ or $p(\bm{x}, \bm{z})$.
\end{itemize} 
See Fig. \ref{WALI} for a visualisation of these networks.

If the two joint distributions are successfully matched, then existing data points can be encoded into latent vectors that follow the same distribution as the sampled input noise. Then, if the latent vectors are passed through the Decoder, the generated images in turn follow the same distribution as the real images. These properties together allow us to manipulate existing data and to interpolate between real data points.

ALI suffers from some limitations, such as training instability and limited ability to faithfully reconstruct images \cite{asveegan17, ALICE}. We find that to successfully train ALI to generate facial images, some tweaks are needed, such as limiting the updates of the Discriminator and ending training before mode collapse occurs. For this reason, we combine the advantages of Wasserstein GANs \cite{arjovsky2017wasserstein,NIPS2017_892c3b1c} with the ALI architecture to improve training stability.

\setlength{\fboxrule}{0.0pt}
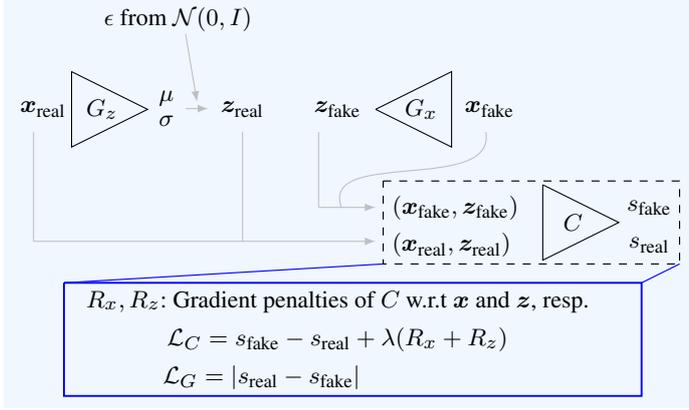
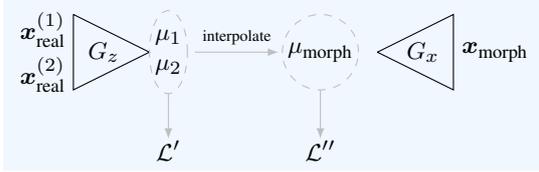
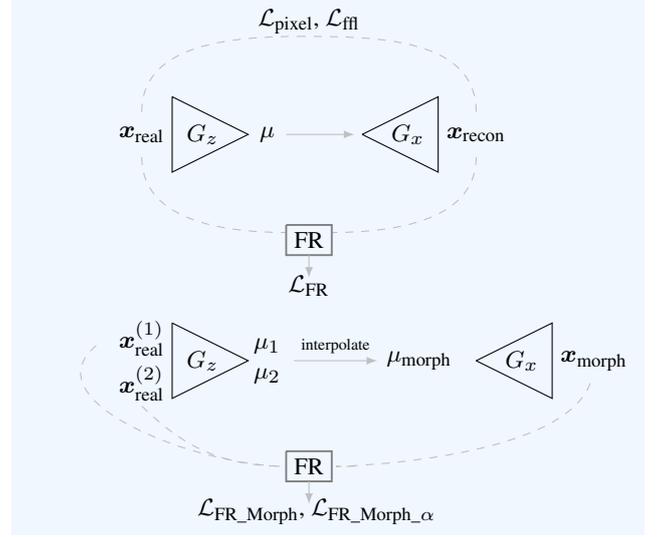
\begin{figure*}[h]
	\begin{subfigure}{0.55\textwidth}
		\centering
		\fcolorbox{black}{aliceblue}{
		\begin{tikzpicture}
		
			\draw[black, -] (0,0.5) -- (0,-0.5) -- (1,0) -- (0,0.5);
			\draw[black, -] (4.0,0) -- (5.0,0.5) -- (5.0,-0.5) -- (4.0,0);
			\draw[black, -] (6.2,-1.0) -- (6.2,-2.0) -- (7.2,-1.5) -- (6.2,-1.0);
			\node[inner sep=0pt] (whitehead) at (0.4,0.0)
			{\text{$G_z$}};
			\node[inner sep=0pt] (whitehead) at (4.6,0.0)
			{\text{$G_x$}};
			\node[inner sep=0pt] (whitehead) at (6.6,-1.5)
			{\text{$C$}};
			
			\node[inner sep=0pt] (whitehead) at (-0.375,0.0)
			{\text{$\bm{x}_{\text{real}}$}};
			\node[inner sep=0pt] (whitehead) at (1.25,0.15)
			{\text{$\mu$}};
			\node[inner sep=0pt] (whitehead) at (1.25,-0.15)
			{\text{$\sigma$}};
			\draw[->,lightgray] (1.5,0.01) -- (1.8,0.01);
			\node[inner sep=0pt] (whitehead) at (2.25,0.0)
			{\text{$\bm{z}_{\text{real}}$}};
			\node[inner sep=0pt] (whitehead) at (1.4,1.2)
			{\text{$\epsilon$ from $\mathcal{N}(0,I)$}};
			\draw[->,lightgray] (1.5,1.0) -- (1.65,0.1);
			
			\node[inner sep=0pt] (whitehead) at (3.5,0.0)
			{\text{$\bm{z}_{\text{fake}}$}};
			\node[inner sep=0pt] (whitehead) at (5.5,0.0)
			{\text{$\bm{x}_{\text{fake}}$}};
			\node[inner sep=0pt] (whitehead) at (5.05,-1.3)
			{\text{$(\bm{x}_{\text{fake}},\bm{z}_{\text{fake}})$}};
			\node[inner sep=0pt] (whitehead) at (5.0,-1.8)
			{\text{$(\bm{x}_{\text{real}},\bm{z}_{\text{real}})$}};
			\node[inner sep=0pt] (whitehead) at (7.6,-1.3)
			{\text{$s_{\text{fake}}$}};
			\node[inner sep=0pt] (whitehead) at (7.6,-1.8)
			{\text{$s_{\text{real}}$}};

			\draw[-, lightgray] (-0.5,-0.3) -- (-0.5,-1.75);
			\draw[-, lightgray] (2.25,-0.3) -- (2.25,-1.75);	
			\draw[->, lightgray] (-0.5,-1.75) -- (4.0,-1.75);
			
			\draw[-, lightgray] (3.25,-0.3) -- (3.25,-1.3);
			\draw[-, lightgray] (5.45,-0.3) to [out=-80,in=110] (3.55,-1.3);	
			\draw[->, lightgray] (3.25,-1.3) -- (4.0,-1.3);	
		\draw[line width=0.15mm, black, dashed] (4.1,-0.95) -- (8.0,-0.95) -- (8.0,-2.05) -- (4.1,-2.05) -- (4.1,-0.95);
		
		\draw[line width=0.25mm, blue] (-0.1,-2.3) -- (7.5,-2.3) -- (7.5,-3.8) -- (-0.1,-3.8) -- (-0.1,-2.3);	
		\draw[line width=0.15mm, blue] (4.1,-2.05) -- (-0.1,-2.3) ;		
		\draw[line width=0.15mm, blue] (8.0,-2.05) -- (7.5,-2.3) ;		
		
		\node[inner sep=0pt] (whitehead) at (3.5,-2.55)
		{\text{$R_x, R_z$: Gradient penalties of $C$ w.r.t $\bm{x}$ and $\bm{z}$, resp.}};
		\node[inner sep=0pt] (whitehead) at (3.5,-3.05)
		{\text{$\mathcal{L}_C = s_\text{fake} - s_\text{real} + \lambda (R_x+R_z)$}};
		\node[inner sep=0pt] (whitehead) at (2.5,-3.55)
		{\text{$\mathcal{L}_G = |s_\text{real} - s_\text{fake}|$}};
		\end{tikzpicture}
		}
		\caption{\label{WALI_losses_baseline}Baseline losses, see Section \ref{baseline}.}
	\end{subfigure}
	\begin{subfigure}{0.4\textwidth}
		\centering
		\fcolorbox{black}{aliceblue}{
		\begin{tikzpicture}
		\newcommand{\shiftdown}{3.0}
		
		\draw[black, -] (0,0.5) -- (0,-0.5) -- (1,0) -- (0,0.5);
		\draw[black, -] (2.5,0) -- (3.5,0.5) -- (3.5,-0.5) -- (2.5,0);

		\node[inner sep=0pt] (whitehead) at (0.4,0.0)
		{\text{$G_z$}};
		\node[inner sep=0pt] (whitehead) at (3.1,0.0)
		{\text{$G_x$}};
		
		\node[inner sep=0pt] (whitehead) at (-0.4,0.0)
		{\text{$\bm{x}_{\text{real}}$}};
		\node[inner sep=0pt] (whitehead) at (1.25,0.0)
		{\text{$\mu$}};
		
		\draw[->,lightgray] (1.5,0.0) -- (2.4,0.0);
		
		\node[inner sep=0pt] (whitehead) at (4.0,0.0)
		{\text{$\bm{x}_{\text{recon}}$}};
		
		\draw[-, lightgray, dashed] (-0.4,0.3) to [out=90,in=180] (1.7,1.3);
		\draw[-, lightgray, dashed] (4.0,0.3) to [out=90,in=0] (1.7,1.3);
		\node[inner sep=0pt] (whitehead) at (1.8,1.5)
		{\text{$\mathcal{L}_{\text{pixel}}$, $\mathcal{L}_{\text{ffl}}$}};
		
		\draw[-, lightgray, dashed] (-0.4,-0.3) to [out=-90,in=180] (1.5,-1.3);
		\draw[-, lightgray, dashed] (4.0,-0.3) to [out=-90,in=0] (2.1,-1.3);
		\draw[line width=0.25mm, gray] (1.5,-1.2) -- (2.1,-1.2) -- (2.1,-1.6) -- (1.5,-1.6) -- (1.5,-1.2);	
		\node[inner sep=0pt] (whitehead) at (1.8,-1.4)
		{\text{FR}};

		\draw[->,lightgray] (1.8,-1.6) -- (1.8,-1.9);
		\node[inner sep=0pt] (whitehead) at (1.8,-2.0)
		{\text{$\mathcal{L}_{\text{FR}}$}};

		
		\draw[black, -] (0,0.5- \shiftdown) -- (0,-0.5- \shiftdown) -- (1,0- \shiftdown) -- (0,0.5- \shiftdown);
		\draw[black, -] (4.0,0- \shiftdown) -- (5.0,0.5- \shiftdown) -- (5.0,-0.5- \shiftdown) -- (4.0,0- \shiftdown);
		\node[inner sep=0pt] (whitehead) at (0.4,0.0- \shiftdown)
		{\text{$G_z$}};
		\node[inner sep=0pt] (whitehead) at (4.6,0.0- \shiftdown)
		{\text{$G_x$}};
		
		\node[inner sep=0pt] (whitehead) at (-0.4,0.3- \shiftdown)
		{\text{$\bm{x}^{(1)}_{\text{real}}$}};
		\node[inner sep=0pt] (whitehead) at (-0.4,-0.3- \shiftdown)
		{\text{$\bm{x}^{(2)}_{\text{real}}$}};
		\node[inner sep=0pt] (whitehead) at (1.25,0.2- \shiftdown)
		{\text{$\mu_1$}};
		\node[inner sep=0pt] (whitehead) at (1.25,-0.2- \shiftdown)
		{\text{$\mu_2$}};
		\node[inner sep=0pt] (whitehead) at (3.25,-0.0- \shiftdown)
		{\text{$\mu_{\text{morph}}$}};

		\node[inner sep=0pt] (whitehead) at (2.15,0.2- \shiftdown)
		{\text{\tiny interpolate}};
		\draw[->,lightgray] (1.6,0.0- \shiftdown) -- (2.7,0.0- \shiftdown);
		
		\node[inner sep=0pt] (whitehead) at (5.55,-0.0- \shiftdown)
		{\text{$\bm{x}_{\text{morph}}$}};
		
		\draw[-, lightgray, dashed] (-1.0,0.2- \shiftdown) to [out=220,in=180] (1.5,-1.4- \shiftdown);
		\draw[-, lightgray, dashed] (-0.4,-0.6- \shiftdown) to [out=-45,in=180] (1.5,-1.4- \shiftdown);
		\draw[-, lightgray, dashed] (5.5,-0.3- \shiftdown) to [out=-110,in=-180] (2.4,-1.4- \shiftdown);
		\draw[line width=0.25mm, gray] (1.5,-1.2- \shiftdown) -- (2.1,-1.2- \shiftdown) -- (2.1,-1.6- \shiftdown) -- (1.5,-1.6- \shiftdown) -- (1.5,-1.2- \shiftdown);	
		\node[inner sep=0pt] (whitehead) at (1.8,-1.4- \shiftdown)
		{\text{FR}};
		
		\draw[->,lightgray] (1.8,-1.6- \shiftdown) -- (1.8,-1.9- \shiftdown);
		\node[inner sep=0pt] (whitehead) at (1.9,-2.0- \shiftdown)
		{\text{$\mathcal{L}_{\text{FR\_Morph}}$, $\mathcal{L}_{\text{FR\_Morph\_}\alpha}$}};
		
		\end{tikzpicture}
		}
		\caption{\label{WALI_losses_finetuning_recon}Losses for finetuning, Section \ref{finetune}.}
	\end{subfigure}
	
	\begin{subfigure}{0.575\textwidth}
		\centering
		\fcolorbox{black}{aliceblue}{
		\begin{tikzpicture}
		\draw[black, -] (0,0.5) -- (0,-0.5) -- (1,0) -- (0,0.5);
		\draw[black, -] (4.0,0) -- (5.0,0.5) -- (5.0,-0.5) -- (4.0,0);
		\node[inner sep=0pt] (whitehead) at (0.4,0.0)
		{\text{$G_z$}};
		\node[inner sep=0pt] (whitehead) at (4.6,0.0)
		{\text{$G_x$}};
		
		\node[inner sep=0pt] (whitehead) at (-0.4,0.3)
		{\text{$\bm{x}^{(1)}_{\text{real}}$}};
		\node[inner sep=0pt] (whitehead) at (-0.4,-0.3)
		{\text{$\bm{x}^{(2)}_{\text{real}}$}};
		\node[inner sep=0pt] (whitehead) at (1.25,0.2)
		{\text{$\mu_1$}};
		\node[inner sep=0pt] (whitehead) at (1.25,-0.2)
		{\text{$\mu_2$}};
		\node[inner sep=0pt] (whitehead) at (3.25,-0.0)
		{\text{$\mu_{\text{morph}}$}};

		\node[inner sep=0pt] (whitehead) at (2.15,0.2)
		{\text{\tiny interpolate}};
		\draw[->,lightgray] (1.6,0.0) -- (2.7,0.0);
		
		\node[inner sep=0pt] (whitehead) at (5.55,-0.0)
		{\text{$\bm{x}_{\text{morph}}$}};
		
		\draw[dashed, lightgray] (1.25cm,0cm) ellipse[x radius=0.25cm,y radius=0.55cm];
		\draw[dashed, lightgray] (3.25cm,0cm) ellipse[x radius=0.5cm,y radius=0.5cm];
		
		\draw[->,lightgray] (1.25,-0.55) -- (1.25,-1.15);
		\draw[->,lightgray] (3.25,-0.5) -- (3.25,-1.15);

		\node[inner sep=0pt] (whitehead) at (1.25,-1.3)
		{\text{$\mathcal{L}'$}};
		\node[inner sep=0pt] (whitehead) at (3.25,-1.3)
		{\text{$\mathcal{L}''$}};

		\end{tikzpicture}
		}
	\caption{\label{WALI_losses_finetuning_morph}Losses from Section \ref{optimisation} are used to optimise the selection of latent embeddings.}
	\end{subfigure}
	\caption{\label{WALI_losses}The losses used in WALI.}
	
\end{figure*}

First, we adapt ALI to include Wasserstein elements and train until convergence, see Fig. \ref{WALI_losses_baseline}. Next, we finetune using losses to encourage the system to generate difficult morphs. We do this using losses on the image level that encourage the system to faithfully reconstruct normal images, but also use a Face Recognition (FR) system to ensure the reconstructed images maintain identity information, see Fig. \ref{WALI_losses_finetuning_recon}. We use the same FR system to nudge the system to generate morphs that approximate worst-case morphs, see Fig. \ref{WALI_losses_finetuning_morph}.

\subsection{Baseline training}\label{baseline}
We mainly follow the ALI training procedure, but replace transposed convolutions with upsampling and size-maintaining convolutions to avoid chequerboard artefacts \cite{odena2016deconvolution}. We also remove the sigmoid output layer in the Discriminator, so that it no longer outputs values between 0 and 1 (where 0 fake, 1 real), but instead outputs a \textit{score}, making the Discriminator network a \textit{Critic} ($C$). A higher Critic score indicates that an image looks more real, and a lower score indicates that according to the Critic the generated image looks more fake. We follow the approach from \cite{NIPS2017_892c3b1c}, i.e. we update $G_z$ and $G_x$ after every fifth update of the Critic. The Critic in turn is trained to output larger scores for real images and vice versa, and to ensure Lipschitz continuity a gradient penalty is added to the Critic loss. Since WALI is trained to match a joint distribution, we include a gradient penalty $R_z$ w.r.t. latent input and a gradient penalty $R_x$ w.r.t. image input. Following recommendations from \cite{NIPS2017_892c3b1c} we set the gradient penalty weight to 10.

We start with a baseline architecture that generates $32 \times 32$ pixel images. The architecture can be changed to generate higher-resolution images by simply adding layers to the three networks. For example, to generate 64 by 64 pixel images, we add one more convolutional layer before the first layer in $G_z$ and $C$, and one more upsampling and convolution after the last layer in $G_x$.

We train $C$ to minimise
\begin{equation}
	\mathcal{L}_C = s_\text{fake} - s_\text{real} + \lambda (R_x+R_z),
\end{equation}
and $G_z$ and $G_x$ to minimise
\begin{equation}
	\mathcal{L}_G = \lvert s_\text{real} - s_\text{fake}  \rvert,
\end{equation}
where
\begin{align}
	s_\text{real} &= \mathop{\mathbb{E}}_{(\bm{x}_\text{real},\bm{z}_\text{real}) \sim q(x,z)}[C(\bm{x}_\text{real}, \bm{z}_\text{real})], \\
	s_\text{fake} &= \mathop{\mathbb{E}}_{(\bm{x}_\text{fake},\bm{z}_\text{fake}) \sim p(x,z)}[C(\bm{x}_\text{fake}, \bm{z}_\text{fake})], \\
	R_x &= \mathop{\mathbb{E}}_{(\tilde{\bm{x}}, \tilde{\bm{z}}) \sim \tilde{p}(x,z)}[(\lVert \nabla_{\tilde{\bm{x}}} C(\tilde{\bm{x}}, \tilde{\bm{z}}) \rVert_2 - 1)^2],  \\
	R_z &= \mathop{\mathbb{E}}_{(\tilde{\bm{x}}, \tilde{\bm{z}}) \sim \tilde{p}(x,z)}[(\lVert \nabla_{\tilde{\bm{z}}} C(\tilde{\bm{x}}, \tilde{\bm{z}}) \rVert_2 - 1)^2].
\end{align}

\subsection{Finetuning for Morph Generation}\label{finetune}
Once the three networks $G_z, G_x$ and $C$ have converged, we fine-tune them using losses that encourage the network to generate morphs that are close to the worst case. We do this using five different losses. The first two losses are a pixel loss $\mathcal{L}_{\text{pixel}}$ and a Focal Frequency Loss (FFL) \cite{jiang2021focal}, both encourage the generator network to reconstruct images on a pixel level. This second loss $\mathcal{L}_{\text{ffl}}$ has the advantage that it forces the generator to reconstruct more challenging frequencies as well as easier frequencies.

Next, we define losses to manipulate identity information in generated images using an FR system. Without loss of generality\footnote{The only requirement is that $d$ is known and differentiable.}, we assume the FR system used compares images using a dissimilarity score function $d$ that calculates the angle between two latent embedding vectors. We denote the mapping used by the FR system to map images onto latent embeddings by $\varphi$. We use three losses to encourage WALI to generate morphs that contain as much as possible relevant identity information. These are $\mathcal{L}_{\text{FR}}$, $\mathcal{L}_{\text{FR\_Morph\_}\alpha}$ and $\mathcal{L}_{\text{FR\_Morph}}$, which are defined as follows:
\begin{equation}
	\mathcal{L}_{\text{FR}} =  \mathop{\mathbb{E}}_{\bm{x}_\text{real} \sim q(\bm{x})} \left[ d(\varphi(\bm{x}_\text{recon}), \varphi(\bm{x}_\text{real})) \right],
\end{equation}
where $\bm{x}_\text{recon} = G_x(G_z(\bm{x}_\text{real}))$.
\begin{equation}
	\mathcal{L}_{\text{FR\_Morph\_}\alpha} =  \mathop{\mathbb{E}}_{(\bm{x}_\text{real},\bm{z}_\text{real}) \sim q(x,z)} \left[ d(\varphi(\bm{x}^{\alpha}_\text{morph}), \bm{y}^*) \right],
\end{equation}
where 
\begin{equation}
	\bm{x}^{\alpha}_\text{morph} = G_x(\alpha \bm{z}_1 + (1-\alpha) \bm{z}_2)
\end{equation}
for $\bm{z}_1=G_z(\bm{x}_1)$ and $\bm{z}_2=G_z(\bm{x}_2)$, and $0\leq\alpha\leq1$. As defined in Eq. \ref{eq:z_wc}, $\bm{y}^*$ is the worst-case embedding given $\bm{y}_1=\varphi(\bm{x}_1)$ and $\bm{y}_2=\varphi(\bm{x}_2)$.
Finally
\begin{equation}
	\mathcal{L}_{\text{FR\_Morph}} = \mathcal{L}_{\text{FR\_Morph\_}\alpha}, \quad \alpha=0.5.
\end{equation}
In principle, $\mathcal{L}_{\text{FR}}$ and $\mathcal{L}_{\text{FR\_Morph}}$ are the same as $\mathcal{L}_{\text{FR\_Morph\_}\alpha}$, just for fixed $\alpha=1$ and $\alpha=0.5$, resp. We find that including these losses specifically instead of simply increasing the weight for $\mathcal{L}_{\text{FR\_Morph\_}\alpha}$ leads to the network being able to generate more challenging morphs when evaluating with FR systems under black-box settings, see Table \ref{tab0opt}.

All five losses are combined in
\begin{align} \label{eq:losses}
	\mathcal{L} = \gamma_1 \mathcal{L}_{\text{pixel}} &+ \gamma_2 \mathcal{L}_{\text{ffl}} + \gamma_3 \mathcal{L}_{\text{FR}} \nonumber\\
	& + \gamma_4 \mathcal{L}_{\text{FR\_Morph}} + \gamma_5 \mathcal{L}_{\text{FR\_Morph\_}\alpha}.
\end{align}
We use MobileFaceNet (MFN) \cite{MFN} to estimate these losses during training, where we intentionally choose a light-weight network in order to reduce GPU memory needed.
We evaluate our generated morphs with seven FR systems that were not used during training: VGG16 \cite{VGG16}, ArcFace (AF) \cite{arcface}, the Inception ResNet-based FaceNet (INC) \cite{FaceNet}, ElasticFace (EF) \cite{ElasticFace}, CurricularFace (CF) \cite{CurricularFace}, PocketNetS-128 (PN) \cite{PocketNet}, Dlib \cite{dlib09}, and a Commercial Off The Shelf (COTS) system.

\begin{algorithm}[h]
	\begin{algorithmic}
		\State $\theta_{D}, \theta_{G_z}, \theta_{G_x} \gets \text{initialise network parameters}$
		\Repeat
		\State $\bm{x_{\text{real}}}^{(1)}, \ldots, \bm{x_{\text{real}}}^{(N)} $
		\Comment{{\footnotesize Draw $N$ samples from the dataset}}
		\State $\bm{z_{\text{fake}}}^{(1)}, \ldots, \bm{z_{\text{fake}}}^{(N)} $
		\Comment{{\footnotesize Draw $N$ random latent emb.}}
		\State $\bm{z_{\text{real}}}^{(i)} = G_z(\bm{x}^{(i)}),		\quad i = 1, .., N$
		\Comment{{\footnotesize Get real embeddings}}
		\State $\bm{x}_{\text{fake}}^{(i)} = G_x(\bm{z_{\text{fake}}}^{(i)}),		\quad i = 1, .., N$
		\Comment{{\footnotesize Generate fake images}}
		\State $s_\text{real} =
		\frac{1}{N} \sum_{i=0}^N C(\bm{x_{\text{real}}}^{(i)}, \bm{z_{\text{real}}}^{(i)})$
		\Comment{{\footnotesize Critic output for real data}}
		\State $s_\text{fake} =
		\frac{1}{N} \sum_{i=0}^N C(\bm{x_{\text{fake}}}^{(i)}, \bm{z_{\text{fake}}}^{(i)})$
		\Comment{{\footnotesize Critic output for fake data}}
		\vspace{0.3cm}
		
		\State $\bm{x_{\text{recon}}}^{(i)} = G_x(\bm{z_{\text{real}}}^{(i)}),		\quad i = 1, .., N$
		\Comment{{\footnotesize Reconstruct real images}}
		\State $\bm{z}_{\alpha,\text{morph}}^{(i)}= \alpha \bm{z_{\text{real}}}^{(i)} + (1-\alpha)\bm{z_{\text{real}}}^{(j)},	\ j = 2, .., N,1$
		\State $\bm{z_{\text{morph}}}^{(i)}= \frac{1}{2} \bm{z_{\text{real}}}^{(i)} + \frac{1}{2} \bm{z_{\text{real}}}^{(j)},	\ j = 2, .., N,1$
		\Comment{{\footnotesize Translate batch to\\
				\hfill get pairs for morphing}}
		\State $\bm{x}_{\alpha,\text{morph}}^{(i)} = G_x(\bm{z}_{\alpha,\text{morph}}^{(i)}),		\quad i = 1, .., N$
		\State $\bm{x}_{\text{morph}}^{(i)} = G_x(\bm{z_{\text{morph}}}^{(i)}),		\quad i = 1, .., N$
		\Comment{{\footnotesize Generate morphs}}
		
		\State $\bm{y}^{(i)}= \varphi(\bm{x}^{(i)}),		\quad i = 1, .., N$
		\Comment{{\footnotesize Get FR embeddings}}
		\State $\bm{y}_\alpha^{*(i)}= \frac{\alpha \bm{y}^{(i)} + (1-\alpha)\bm{y}^{(j)}}{||\alpha \bm{y}^{(i)} + (1-\alpha)\bm{y}^{(j)}||},	\ j = 2, .., N,1$
		\State $\bm{y}^{*(i)}= \frac{\bm{y}^{(i)}+\bm{y}^{(j)}}{||\bm{y}^{(i)}+\bm{y}^{(j)}||},	\ j = 2, .., N,1$
		\Comment{{\footnotesize Get worst-case emb.}}
		\vspace{0.3cm}
		
		\State $\mathcal{L}_{\text{pixel}} = \frac{1}{N} \sum_{i=0}^N \text{MSE}(\bm{x}^{(i)}, \bm{x}_{\text{recon}}^{(i)})$
		\Comment{{\footnotesize Compute pixel loss}}
		\State $\mathcal{L}_{\text{ffl}} = \frac{1}{N} \sum_{i=0}^N \text{FFL}(\bm{x}^{(i)}, \bm{x}_{\text{recon}}^{(i)})$
		\Comment{{\footnotesize Compute FFL loss}}
		\\
		\State $\mathcal{L}_\text{FR} = \frac{1}{N} \sum_{i=0}^N (\bm{y}^{(i)}, \bm{y}_{\text{recon}}^{(i)})$\\
		\State $\mathcal{L}_{\text{FR\_Morph\_}\alpha} = \frac{1}{N} \sum_{i=0}^N (\bm{y}^{(i)}_{\alpha,\text{morph}}, \bm{y}_\alpha^{*(i)})$\\
		\State $\mathcal{L}_{\text{FR\_Morph\_}} = \frac{1}{N} \sum_{i=0}^N (\bm{y}^{(i)}_{\text{morph}}, \bm{y}^{*(i)})$\\
		\Comment{{\footnotesize Compute FR-based losses}}
		\\
		\State $\mathcal{L}_C = s_\text{fake} - s_\text{real} + \lambda (R_x+R_z)$
		\Comment{{\footnotesize Compute Critic loss}}
		\State $\mathcal{L}_G = |s_\text{fake} - s_\text{real}| + \gamma_1 \mathcal{L}_{\text{pixel}} + \gamma_2 \mathcal{L}_{\text{ffl}} + \gamma_3 \mathcal{L}_{\text{FR}}$
		\State $\hspace{2.7cm} + \gamma_4 \mathcal{L}_{\text{FR\_Morph\_}\alpha} + \gamma_5 \mathcal{L}_{\text{FR\_Morph\_}\alpha}$
		\\
		\Comment{{\footnotesize Compute Generator loss}}
		\State $\theta_{C} \gets \theta_{C} - \nabla_{\theta_{C}} \mathcal{L}_C$
		\Comment{{\footnotesize Gradient update on Critic}}	
		\State $\theta_{G_z} \gets \theta_{G_z} - \nabla_{\theta_{G_z}} \mathcal{L}_G$
		\Comment{{\footnotesize Gradient update on Encoder}}	
		\State $\theta_{G_x} \gets \theta_{G_x} - \nabla_{\theta_{G_x}} \mathcal{L}_G$
		\Comment{{\footnotesize Gradient update on Decoder}}		
		\Until{convergence}
	\end{algorithmic}
	\caption{\label{alg:pseudo} Our training procedure.}
\end{algorithm}

\subsection{Optimisation}\label{optimisation}
After training the three networks we freeze their weights and optimise the selection of latent embeddings. For each pair of images for morphing, we apply optimisation to select good initial embeddings, and then optimise again to find an embedding that when passed through $G_x$ leads to a morph that is close to the worst case. We use an Adam optimiser \cite{ADAM} with hyperparameters $\alpha=0.05$, $\beta_1=0.9, \beta_2= 0.999$.\\
\\
\noindent \textbf{Optimisation phase 1}: \\
Start with  $\bm{z}_1=G_z(\bm{x})$ and optimise $\bm{z}_1$ using 
\begin{equation}\label{L1}
	\mathcal{L}'=||\bm{x}_1 - G_x(\bm{z}_1)||^2 + ||\varphi(\bm{x}_1) - \varphi(G_x(\bm{z}_1))||^2.
\end{equation}
Do the same to optimise the selection of $\bm{z}_2$.
\\
\\
\textbf{Optimisation phase 2}: \\
Start with $\bm{z}_\text{morph}=\frac{\bm{z}_1+\bm{z}_2}{2}$ and optimise using 
\begin{equation}\label{L2}
	\mathcal{L}''=||\bm{y}^* - \varphi(G_x(\bm{z}_\text{morph}))||^2.
\end{equation}

In both phases a second FR system can be included to improve the effects of optimisation. Let $\varphi_2$ be the mapping corresponding to the second FR system. Then Equations \ref{L1}, \ref{L2} are extended by adding $||\varphi_2(\bm{x}_1) - \varphi_2(G_x(\bm{z}_1))||^2$ to $\mathcal{L}'$ and $||\bm{y}^{**} - \varphi_2(G_x(\bm{z}_\text{morph}))||^2$ to $\mathcal{L}''$, where $\bm{y}^{**}$ is the worst-case embedding in the second FR system's latent space. In both phases the second FR system can be given more or less weight by weighting the new summands.

\subsection{Optimisation with pretrained Generator}
All three WALI networks can be trained simultaneously and from scratch. Alternatively, with very little adaptation our two-phase optimisation approach also allows the use of an existing Encoder and Generator. We apply optimisation guided by MFN and EF in two phases, similar to Optimisation Phase 1 and 2, when using an existing StyleGAN Generator and an Encoder provided by \cite{encoder4editing}. This Encoder was trained to invert the StyleGAN Generator mapping, but unlike WALI, the Generator and Encoder were not trained simultaneously. Since this approach is similar to MIPGAN~\cite{MIPGAN}, we call it \emph{improved MIPGAN}.

\section{Experiments} \label{Experiments} 

We report results for experiments with WALI models that generate 128$\times$128 images, since training time and GPU memory requirements increase significantly when (1) including FR losses during training and (2) increasing the size of the model to generate higher-resolution images.
We report our results on 128$\times$128 images, but have successfully managed to generate visually convincing images up to dimensions of $d_1=512\times512\times 3$ (compared to $d_1 = 64\times 64\times 3$ for ALI images), see Figure \ref{fig:WALI512}.

We train the three WALI networks without any losses other than the Wasserstein and gradient penalty loss until they converge, which takes about 400 epochs with a batch size of 32. We then finetune the networks by adding the losses in Eq. \ref{eq:losses} and training for another 85 epochs. Our model was implemented with Pytorch \cite{PyTorch}, training and testing experiments were conducted on a computer equipped with two Nvidia GeForce Titan-X GPUs (12G).

We use MobileFaceNet \cite{MobileFaceNet} to implement the loss functions in Eq. \ref{eq:losses} during training. To guide optimisation towards an embedding that is ``close'' to the worst case we also use MobileFaceNet. Additionally, we report results for experiments in which we used two FR systems during optimisation. We also apply our two-phase optimisation approach using a StyleGAN Generator and an Encoder network from \cite{encoder4editing}, which we call ``improved MIPGAN''.

\subsection{Datasets} \label{dataset}
We use a dataset of in total 21,772 facial images from the FRGC dataset \cite{FRGC} and separate them into 18,143 training and 3,629 validation images, with no overlap in identities. We add 32,869 images with frontal pose from FFHQ \cite{StyleGAN_FlickR} to the training set. Training without including FFHQ images in the training set was also successful, but including FFHQ improves the results, especially when evaluating with FR (as opposed to only by visual inspection).

We create four sets of morphs using the validation set: landmark-based morphs, GAN-based morphs using \mbox{MIPGAN-I}~\cite{MIPGAN}, approximations of worst-case morphs generated by our WALI method, and approximations of worst-case morphs generated using our improved MIPGAN implementation. 
We select 75 pairs of similar identities by calculating a mean FR embedding for each identity: $\overline{z} = \frac{1}{n} \sum_i^n \varphi(x_i)$, and then selecting those pairs for which the mean FR embeddings are most similar. For each pair of identities we select all faces with neutral expression and from all possible combinations we randomly select 506 image pairs for morphing.

For each pair $(x_1,x_2)$ we create three landmark morphs, one MIPGAN morph, one WALI worst-case approximation for each FR system used for optimisation (seven in total), and one improved MIPGAN worst-case approximation, see Fig. \ref{fig:worst-case}. The three landmark morphs comprise one full morph - the faces and also the background of both original images are morphed - and two spliced morphs - full morphs spliced into the background of each of the original images respectively to remove ghosting artefacts. After freezing WALI's weights, a worst-case approximation is generated by applying 150 optimisation steps in phase 1 and 150 steps in phase 2 (Section \ref{optimisation}). For our improved MIPGAN morphs we also apply 150 optimisation steps in two phases, this time using a StyleGAN Generator and Encoder. We notice that there is a difference in behaviour between the newer FR systems ElasticFace and CurricularFace compared to the other FR systems we use for optimisation, which we describe in Section \ref{Results}. For this reason, whenever we optimise with MobileFaceNet and one of these two FR systems, we weight the losses corresponding to the latter with 2. In all other cases the optimisation losses as defined in Section \ref{optimisation} are weighted equally. We did not extensively analyse the effect of weighting losses differently, so in other applications this may need to be examined further in order to select weights that suitably balance the different losses. For image generation tools and/or MAD methods, we are aware that it would be better to use datasets that are more balanced and include more variation in terms of gender, age, ethnicity and we encourage the research community to take this into consideration for future research.

We also compare different GAN- and landmark-based morphs \cite{9746477, Sarkar2020} created using images from the FRLL \cite{debruine_jones_2017} dataset. The FRLL dataset consists of 102 identities and two images per identity. For each identity one image with neutral expression is provided that is suitable for morph generation. Five morph datasets are provided: WebMorph, OpenCV, FaceMorpher, and AMSL are landmark-based morphs, StyleGAN morphs are GAN-based morphs. AMSL consists of 2175 landmark-based morphs. When using a landmark-based morphing tool, a morph based on two images can be spliced into the background of either the first or the second image. Since in the AMSL dataset both options are not always provided, we only evaluate with identity pairs for which both spliced morphs are provided. We do this to enable a fair comparison of all morphing methods. WebMorph, OpenCV and FaceMorpher consist of full morphs only, i.e. they contain obvious morphing artefacts.

\section{Evaluation Metrics} \label{metrics}
To measure and compare the performance of our model, we calculate the Morphing Attack Potential (MAP) \cite{MorphingAttackPotential} for $r=1$ verification attempt and $c=1$ FR system, which is the same as the Mated Morph Presentation Match Rate, abbreviated MMPMR($t$). We consider morphs based on two identities, in which case the MMPMR($t$) \cite{SNR17} is the proportion of (morphing) attacks for which both contributing identities are considered a match by the FR system when using a threshold~$t$.
\begin{equation}
\text{MMPMR}(t)=\frac{1}{M}\sum_{m=1}^M \left\{ \max(d_1,d_2) <t \right\}
\end{equation}
where $d_1$ and $d_2$ are the dissimilarity scores between the morph and a probe image of the first and second identity, respectively. $M$ is the number of morphed images.

We report MMPMR values for nine different FR systems. For each FR system we set $t$ such that the false non-match rate is minimal while the false match rate$<$0.1\%. Higher MMPMR values indicate higher vulnerability to morphing attacks. It would be possible to compute MAP values for $c>1$, but for WALI morphs we always treat one or two FR systems as white-box systems, so this might lead to unfair comparisons. Instead, we would have to compute different MAP matrices for all morphing techniques excluding one or two FR systems at a time, which would become very messy. For this reason we choose to only report the MMPMR.

\subsection{Morphing Attack Detection}\label{MAD}
We evaluate morphs generated using WALI with two MAD methods. The first is a single image-based morphing (S-MAD) approach, based on Support Vector Machine (SVM) trained with Local Binary Pattern (LBP) features, that learns to detect morphed images based on image texture described using LBP features \cite{MAD,robustMAD}. The second is a differential image-based (D-MAD) method that is based on Deep Learning (DL) features \cite{scherhag2020deep}. 

We train both MAD methods using the FRGC images we also used to train WALI. While the LBP-based approach can successfully detect WALI morphs, this may simply be due to the similarity of WALI morphs to the FRGC training data. To show that this is the case and that it is insufficient to train with landmark-based morphs only, we also train the LBP approach using FRLL and AMSL. We include 20\% of the landmark-based morphs (selected randomly) in the training set, due to the class imbalance. Because of the low number of genuine pairs (only one pair per identity) we do not train the D-MAD approach with this dataset. \\

We report the performance of these two MAD methods using the Bona fide Presentation Classification Error Rate (BPCER): the proportion of bona fide images that are incorrectly labelled as morphs and the Attack Presentation Classification Error Rate (APCER): the proportion of (morphing) attacks that are misidentified as bona fides. Higher values of BPCER and APCER indicate higher vulnerability of an MAD system to morphing attacks.

\newcommand{\wi}{.08\linewidth}
\newcommand{\gap}{\hspace{-0.25cm}}
\newcommand{\inc}{\includegraphics[width=\wi,valign=m]}
\newcommand{\gr}{\cellcolor{gray!25}}
\begin{figure*}[h]
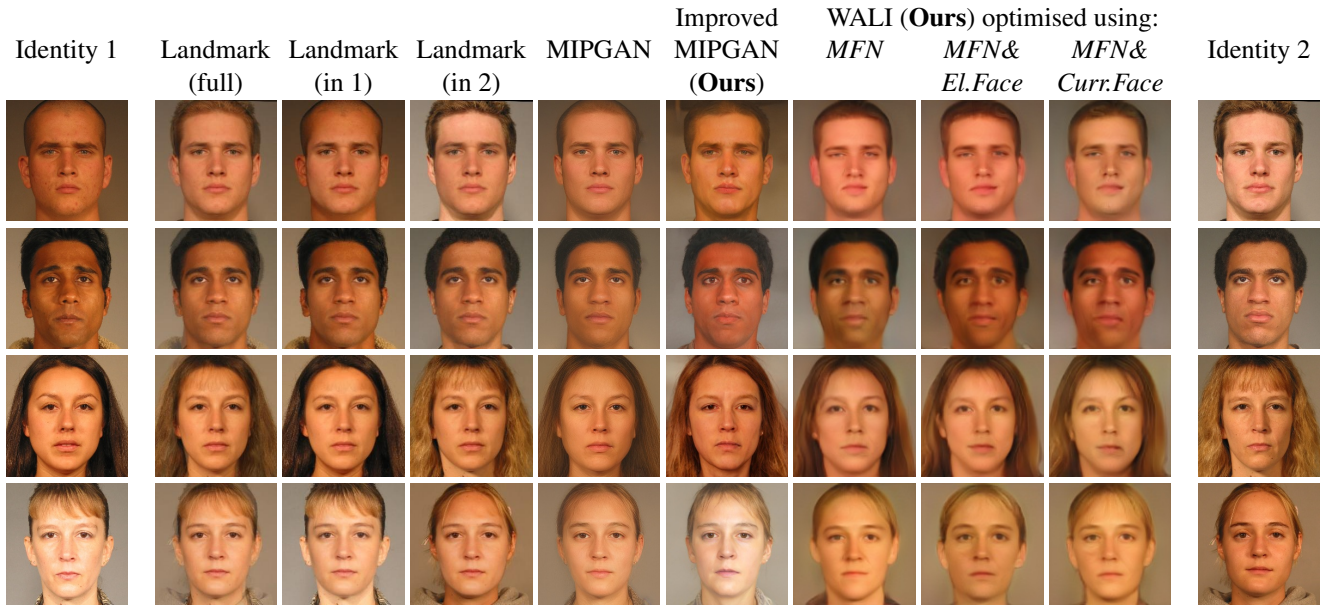

	\centering
	\small
	\resizebox{.99\textwidth}{!}{
		\begin{tabular}{cccccccccc}
			&		 			&				 		&				 		&			 			& \gap Improved 		& \multicolumn{3}{c}{WALI ({\bf{Ours}}) optimised using:}				 		&				\\
			Identity 1			& Landmark 			& \gap Landmark			& \gap Landmark			& \gap MIPGAN			& \gap MIPGAN			& \gap \it{MFN}			& \gap \it{MFN\&}		& \gap \it{MFN\&}		& Identity 2	\\
			& (full) 			& \gap (in 1) 			& \gap (in 2)  			& 			 			& \gap ({\bf{Ours}})			& \gap 					& \gap \it{El.Face}		& \gap \it{Curr.Face}	&				\\
			
			\inc{04361d171}	&\inc{04361d171_04484d36_50}	&\gap\inc{04361d171_04484d36_50_in0}	&\gap\inc{04361d171_04484d36_50_in1}	&\gap\inc{04361d171_04484d36_mipgan}	&\gap\inc{04361d171_04484d36_improvedMIPGAN}	&\gap\inc{04361d171_04484d36_WALI}	&\gap\inc{04361d171_04484d36_WALI_ef}	&\gap\inc{04361d171_04484d36_WALI_cf}	& \inc{04484d36}\\
			\vspace{-0.3cm}\\
			
			\inc{04395d210}	&\inc{04395d210_04779d14_50}	&\gap\inc{04395d210_04779d14_50_in0}	&\gap\inc{04395d210_04779d14_50_in1}	&\gap\inc{04395d210_04779d14_mipgan}	&\gap\inc{04395d210_04779d14_improvedMIPGAN}	&\gap\inc{04395d210_04779d14_WALI}	&\gap\inc{04395d210_04779d14_WALI_ef}	&\gap\inc{04395d210_04779d14_WALI_cf}	& \inc{04779d14}\\
			\vspace{-0.3cm}\\
			
			\inc{04316d133}	&\inc{04316d133_04418d18_50}	&\gap\inc{04316d133_04418d18_50_in0}	&\gap\inc{04316d133_04418d18_50_in1}	&\gap\inc{04316d133_04418d18_mipgan}	&\gap\inc{04316d133_04418d18_improvedMIPGAN}	&\gap\inc{04316d133_04418d18_WALI}	&\gap\inc{04316d133_04418d18_WALI_ef}	&\gap\inc{04316d133_04418d18_WALI_cf}	& \inc{04418d18}\\
			\vspace{-0.3cm}\\
			
			\inc{04418d251}	&\inc{04418d251_04495d02_50}	&\gap\inc{04418d251_04495d02_50_in0}	&\gap\inc{04418d251_04495d02_50_in1}	&\gap\inc{04418d251_04495d02_mipgan}	&\gap\inc{04418d251_04495d02_improvedMIPGAN}	&\gap\inc{04418d251_04495d02_WALI}	&\gap\inc{04418d251_04495d02_WALI_ef}	&\gap\inc{04418d251_04495d02_WALI_cf}	& \inc{04495d02}\\
			
		\end{tabular}
	}
	\caption{\label{fig:worst-case}Examples of landmark-based and different GAN-based morphs based on FRGC images.}
\end{figure*}

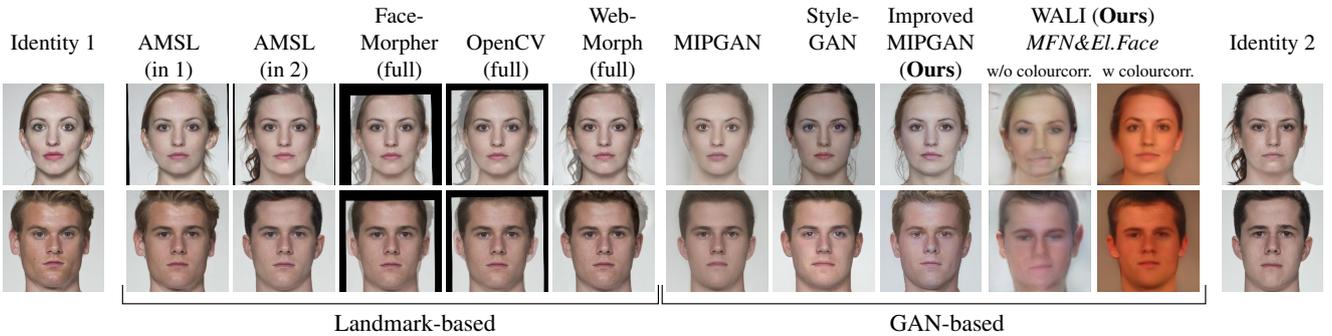
\begin{figure*}[h]
	\centering
	\small
	\resizebox{.99\textwidth}{!}{
		\begin{tabular}{ccccccccccccc}
			&		 			&				 	& Face-		&			& Web-	 		&			 				& Style-		& \gap Improved 		& \multicolumn{2}{c}{\gap WALI ({\bf{Ours}})} 											&				\\
			Identity 1	& \gap AMSL			& \gap AMSL			& Morpher 	& OpenCV	& Morph	 		& \hspace{-0.15cm}MIPGAN	& GAN			& \gap MIPGAN			& \multicolumn{2}{c}{\gap \it{MFN\&El.Face}}											& Identity 2	\\
			& \gap (in 1) 		& \gap (in 2)  		& (full) 	& (full)	& (full)		& 			 				&				& \gap ({\bf{Ours}})			& \gap{\scriptsize w/o colourcorr.}			& \gap{\scriptsize w colourcorr.}	&				\\
			\inc{002_03}		&\inc{002_091_AMSL}	&\gap\inc{091_002_AMSL}	&\gap\inc{002_091_facemorpher}	&\gap\inc{002_091_opencv} &\gap\inc{002_091_webmorph}	&\hspace{-0.15cm}\inc{002_091_MIPGAN}	&\gap\inc{002_091_stylegan}	&\gap\inc{002_091_imprMIPGAN}		&\gap\inc{002_091_WALI}		&\gap\inc{002_091_WALI_colorcorr}	& \inc{091_03}\\
			\vspace{-0.3cm}\\
			\inc{012_03}		&\inc{012_063_AMSL}	&\gap\inc{063_012_AMSL}	&\gap\inc{012_063_facemorpher}	&\gap\inc{012_063_opencv} &\gap\inc{012_063_webmorph}	&\hspace{-0.15cm}\inc{012_063_MIPGAN}	&\gap\inc{012_063_stylegan}	&\gap\inc{012_063_imprMIPGAN}		&\gap\inc{012_063_WALI}		&\gap\inc{012_063_WALI_colorcorr}	& \inc{063_03}\\
		\end{tabular}
	}
	\begin{tikzpicture}[overlay,remember picture]
	\draw[line width=0.15mm, black] (-16.1,-1.75) -- (-16.1,-2.0) -- (-9.05,-2.0) -- (-9.05,-1.75);
	\draw[line width=0.15mm, black] (-9.0,-1.75) -- (-9.0,-2.0) -- (-1.85,-2.0) -- (-1.85,-1.75);
	\node[inner sep=0pt] (whitehead) at (-12.25,-2.25) {\text{Landmark-based}};
	\node[inner sep=0pt] (whitehead) at (-5.25,-2.25)	{\text{GAN-based}};
	\end{tikzpicture}
	\vspace{0.5cm}
	\caption{\label{fig:worst-case-frll}Examples of morphs based on FRLL images. WALI (and other morph methods) are trained on another dataset and applied to FRLL images, which have different lighting and colour balance. WALI may not generalise well to unseen data, mainly because of the simple WALI generator which cannot compete with more powerful GANs. Incorporating StyleGAN in our WALI pipeline results in `Improved MIPGAN', giving visually convincing results. }
\end{figure*}

\newcommand{\red}{\cellcolor{red!25}}
\newcommand{\tkzmI}{\tikzmark{I}}
\newcommand{\tkzmJ}{\tikzmark{J}}
\newcommand{\tkzmK}{\tikzmark{K}}
\newcommand{\tkzmL}{\tikzmark{L}}
\newcommand{\multisev}{\multicolumn{7}}
\newcommand{\multi}{\multicolumn{1}}

\begin{table*}[t]
	\centering
	\caption{MMPMR values for landmark- and GAN-based morphs. The second-to-last column shows the theoretical worst case for each respective FR system. Grey cells indicate evaluation was under white-box assumptions, i.e. this FR system was used during optimisation. The more challenging the morphs, the higher the MMPMR. To show that there is a trade-off between FR performance and vulnerability to morphing attacks we report the False Non-Match Rate (FNMR) (\%) at which the False Match Rate $<0.1\%$ in the last column. The morphing methods highlighted in blue are closest to the worst case for almost all FR systems.}\label{tab1}
	\resizebox{0.99\textwidth}{!}{
		\begin{tabular}{|l|l|l|l|l|l|l|l|l|l|l||l||l|}
			\hline
			&		&			&	Improved	& \multisev{c||}{ WALI (Ours) with optimisation using:}	& & \\
			
			& Land-	& MIPGAN	&	MIPGAN		& \multi{c}{ \ \it{MFN} }
			&\multi{c}{\it{MFN}\&} 
			&\multi{c}{\it{MFN}\&}
			&\multi{c}{\it{MFN}\&}
			&\multi{c}{\it{MFN}\&}
			&\multi{c}{\it{MFN}\&}
			&\multi{c||}{\it{MFN}\&} 	& Worst		& FNMR\\
			& mark	&			&	(Ours)		& \multi{c}{}	&\multi{c}{\it{El.Face}} 
			&\multi{c}{\it{Curr.Face}}
			&\multi{c}{\it{ArcFace}}
			&\multi{c}{\it{Inception}}
			&\multi{c}{\it{PocketNet}} 
			&\multi{c||}{\it{VGG16}} & Case	&\\	
			\hline
			\textit{MobileFaceNet}
			& 65.7		& 71.9	& \tkzmI\gr91.8	&\gr 96.8	&\tkzmK\gr96.6	& \gr96.6	& \gr96.8 	& \gr97.0 	& \gr97.0	& \gr96.6	& 97.5 	& 0.5\\
			\hline
			\textit{ElasticFace}
			& 56.9		& 18.9	& \gr83.0		& 14.0		& \gr81.8		& 60.4		& 25.7		& 20.8 		& 18.7		& 18.1		& 98.8	& 0.0 \\
			\hline
			\textit{CurricularFace}
			& 45.9		& 11.1	& 60.9			& 8.6 		& 46.6			& \gr68.3	& 14.8		& 12.2		& 10.5		& 13.2		& 99.0	& 0.0 \\
			\hline
			\textit{ArcFace}
			& 70.7		&\gr62.9& 84.5			& 64.6 		& 76.2			& 75.6		& \gr90.4	& 71.7		& 70.2		& 70.2		& 97.9	& 0.2\\
			\hline
			\textit{Inception}	
			& 36.8		& 37.0	& 51.1			& 37.6 		& 47.2			& 46.4		& 43.7		& \gr58.0	& 41.2		& 42.5		& 71.8	& 3.4\\
			\hline
			\textit{PocketNet}
			& 34.1		& 34.2	& 49.0			& 48.0 		& 49.3			& 48.7		& 51.4		& 51.3		& \gr63.5	& 50.3		& 84.2	& 3.8 \\
			\hline
			\textit{VGG16}
			& 36.4		& 32.7	& 42.1			& 35.4 		& 39.4			& 40.1		& 40.1		& 41.1		& 38.5		& \gr56.2	& 92.0	& 7.6\\
			\hline
			\textit{Dlib}
			& 45.1		& 37.2	& 42.4			& 27.3 		& 32.6			& 31.4		& 32.5		& 32.9		& 30.0		& 32.2		& 72.3 	& 5.8\\
			\hline
			\textit{COTS}
			& 99.8		& 93.4	& \tkzmJ98.6	& 71.4 		& 94.6			&\tkzmL95.5	& 79.6		& 80.4		& 76.3 		& 75.0		& n/a	& 0.0\\
			\hline
		\end{tabular}
		\begin{tikzpicture}[overlay,remember picture]
		\draw[blue] ([shift={(-1.25ex,1.9ex)}]pic cs:I) rectangle ([shift={(10.0ex,-0.7ex)}]pic cs:J);
		\draw[blue] ([shift={(-1.25ex,1.9ex)}]pic cs:K) rectangle ([shift={(10.0ex,-0.7ex)}]pic cs:L);
		\end{tikzpicture}
	}
\end{table*}

\begin{table*}[t]
	\caption{MMPMR for WALI morphs without any optimisation steps. The more challenging the morphs, the higher the MMPMR.}\label{tab0opt}
	\centering
	\resizebox{0.8\textwidth}{!}{
		\begin{tabular}{|l|l|l||l|l|l|l|}
			\hline
			& MIPGAN	& WALI (Ours) 			& WALI without	& WALI w/o	& WALI w/o	& WALI w/o	\\
			& 			& with all FR losses	& FR losses		& $\mathcal{L}_{\text{FR\_Morph\_}\alpha}$
			& $\mathcal{L}_{\text{FR\_Morph}}$	& $\mathcal{L}_{\text{FR}}$ \\		
			\hline
			\textit{MobileFaceNet}
			& 0.9		& 19.2					& 0.3			& \bf{19.6}	& 19.3		& 14.4		\\
			\hline
			\textit{ElasticFace}	(black box)	
			& 0.0		& 0.0					& 0.0			& 0.0		& 0.0		& 0.0		\\
			\hline
			\textit{Curricularface}	(black box)	
			& 0.0		& 0.0					& 0.0			& 0.0		& 0.0		& 0.0		\\
			\hline
			\textit{ArcFace}	(black box)
			& 0.7		& \bf{13.0}				& 0.1			& 12.0		& 6.9		& 6.6		\\
			\hline
			\textit{Inception}	(black box)	
			& 0.5		& \bf{13.2}				& 0.6			& 7.9		& 9.5		& 9.0		\\
			\hline
			\textit{PocketNet}	(black box)	
			& 1.5		& \bf{18.0}				& 1.1 			& 16.8 		& 17.7 		& 14.0 		\\
			\hline
			\textit{VGG16}	(black box)
			& 2.0		& \bf{10.7}				& 0.4			& 9.5		& 8.0		& 6.0		\\
			\hline
			\textit{Dlib}	(black box)	
			& 5.6		& \bf{10.1}				& 0.3			& 8.1		& 3.7		& 7.5		\\
			\hline
			\textit{COTS}	(black box)	
			& 0.0		& 0.1 					& 0.0 			& \bf{1.5}	& \bf{1.5}	& 1.4 		\\
			\hline
		\end{tabular}
	}
\end{table*}

\newcommand{\tkzmM}{\tikzmark{M}}
\newcommand{\tkzmN}{\tikzmark{N}}
\begin{table*}[t]
	\caption{MMPMR for FRLL morphs. The more challenging the morphs, the higher the MMPMR. Grey cells indicate the FR system was used for optimisation. Of all GAN-based approaches the Improved MIPGAN approach (highlighted in blue) is closest to the worst case for almost all FR systems.}\label{tab_frll}
	\centering
	\resizebox{0.99\textwidth}{!}{
		\begin{tabular}{|l|l|l|l|l||l|l|l|l|l||l|}
			\hline
			& \multicolumn{4}{c||}{ Landmark-based morphing:} & \multicolumn{5}{c||}{ GAN-based morphing:} & 	\\
			\cmidrule{2-10}
			& AMSL		& Face-		&OpenCV	& WebMorph	& Style-& MIPGAN	& Improved		&WALI (Ours)	&WALI (Ours)& Worst	\\
			&			& Morpher	& 		& 			& GAN	& 			& MIPGAN		& MFN\&EF		& MFN\&EF	& Case	\\
			&			&			&		&			&		&			& (Ours)		&w/o colourcorr.	&w colourcorr.&\\
			\hline
			\textit{MobileFaceNet}
			& 64.7		& 89.2		& 86.0	& 90.3		& 22.1	& 74.4		& \tkzmM\gr96.2	& \gr96.7		& \gr96.8	& 99.5	\\
			\hline
			\textit{ElasticFace}	
			& 38.8		& 58.8		& 60.4	& 60.0		& 0.0	& 4.7		& \gr74.4		& \gr26.1		& \gr44.0	& 99.8	\\
			\hline
			\textit{Curricularface}	
			& 32.7		& 53.4		& 55.6	& 54.3		& 0.0	& 2.6		& 44.3			& 3.0			& 11.0		& 99.6	\\
			\hline
			\textit{ArcFace}
			& 58.8		& 67.2		& 64.4	& 65.3		& 1.2	& \gr20.3	& 64.6			& 7.8			& 17.0		& 99.4	\\
			\hline
			\textit{Inception}	
			& 9.6		& 14.2		& 14.2	& 17.3		& 0.3	& 5.0		& 11.2			& 1.2			& 2.4		& 49.3	\\
			\hline
			\textit{PocketNet}	
			& 51.7		& 78.0		& 80.3	& 84.7		& 20.0	& 55.9		& 75.8			& 14.1			& 44.5		& 98.8	\\
			\hline
			\textit{VGG16}
			& 12.1		& 12.8		& 14.2	& 25.2		& 0.4	& 6.8		& 20.5			& 2.3			& 4.7		& 100.0	\\
			\hline
			\textit{Dlib}
			& 0.9		& 1.0		& 1.0	& 0.9		& 0.0	& 0.0		& 0.6			& 0.1			& 0.1		& 16.5	\\
			\hline
			\textit{COTS}	
			& 97.6		& 100.0		& 100.0	& 100.0		& 0.8	& 70.4		& \tkzmN96.6	& 46.6	& 64.7		& n/a	\\
			\hline
			
		\end{tabular}
		\begin{tikzpicture}[overlay,remember picture]
		\draw[blue] ([shift={(-1.25ex,1.9ex)}]pic cs:M) rectangle ([shift={(10.0ex,-0.7ex)}]pic cs:N);
		\end{tikzpicture}
	}
\end{table*}

\section{Results} \label{Results} 

In Fig. \ref{fig:worst-case} we show examples of morphs generated using WALI and compare them with landmark, MIPGAN and improved MIPGAN morphs. WALI morphs are more blurry compared to MIPGAN morphs, which to a large extent is due to MIPGAN relying on a StyleGAN model that generates $1024\times1024$ images while the WALI morphs are $128\times128$-pixel images. In Fig. \ref{fig:WALI512} we show that the visual quality (from a human perspective) can be improved simply by increasing the WALI model size.

We report MMPMR values for one FR system at a time for the case where optimisation was guided by MFN only and for the case where optimisation was guided by two FR systems (MFN+EF, MFN+CF, MFN+AF, MFN+INC, MFN+PN, MFN+VGG), see Table~\ref{tab1}. Dlib is not available as a Pytorch implementation, so we did not optimise using this FR system. When WALI is optimised with two FR systems, the resulting morphs are more challenging than either landmark or MIPGAN morphs for both FR systems used for optimisation. There is an interesting difference in behaviour that sets apart ElasticFace and CurricularFace from other FR systems. Comparing WALI morphs optimised with MFN+AF, MFN+INC, MFN+PN, MFN+VGG to landmark- and MIPGAN- morphs, we see that the MMPMR is closer to the worst case for all black-box tested FR systems except ElasticFace, CurricularFace, Dlib and COTS. At first glance this could be interpreted to mean that ElasticFace and CurricularFace are generally less vulnerable to GAN-based morphing attacks. However, when WALI morphs are optimised using ElasticFace, the resulting morphs are also closer to the worst case when evaluating with CurricularFace and vice versa. When either of the two is used for optimisation, they are no less vulnerable to GAN-based morphing attacks than other FR systems. Interestingly, Dlib - and to a lesser extent also the COTS FR system - is less vulnerable to MIPGAN and WALI morphs than to landmark morphs. It is also interesting to highlight the inverse relationship between performance on normal images and vulnerability to morphing attacks. Comparing the last two columns illustrates this in theory: the two FR systems with the lowest FNMR also have the highest worst-case MMPMR. In practice the same pattern is shown: the FR systems with lower FNMR are indeed more vulnerable to landmark, MIPGAN and WALI morphs.

Table \ref{tab0opt} reports the MMPMR and confirms our hypothesis, that explicitly considering the goal of morphing \textit{during} training leads to more challenging morphs. There may be some amount of trade-off between the two goals when using WALI: generating visually convincing images versus successfully manipulating identity information.

The following four aspects lead to more challenging morphs:
\begin{enumerate}
	\item defining a worst-case embedding that we can use to define losses during training and optimisation,
	\item explicitly training the model to generate morphs,
	\item improving optimisation by splitting it into two phases: before we generate morphs, we select good initial embeddings for each input image,
	\item optimising with more than one FR system.
\end{enumerate}

WALI does not seem to generalise well to other datasets. This can be seen in Table \ref{tab_frll} and Fig. \ref{fig:worst-case-frll}. This is to a large extent due to our WALI Generator (7.8 million parameters) not being able to compete with a more powerful Generator such as StyleGAN (28.3 million parameters). When applying a colour correction to FRLL images so that they more closely resemble FRGC images, the MMPMR of WALI morphs significantly increases, for example from 30.0\% to 11.0\% for CurricularFace or 14.1\% to 44.5\% for PocketNet, indicating that the lower performance of WALI is to a large extent due to the different type of data. In order to illustrate the effect of approximating a worst case when considering FRLL data, we can apply three of the four improvements listed above to existing generative methods. We show that combining a more powerful StyleGAN Generator with the improved optimisation approach in two phases, as well as optimising with two FR systems still leads to closer approximations of worst-case morphs. Morphs generated with our improved MIPGAN implementation have higher MMPMR values than all other GAN-based morphs, and also higher MMPMR than AMSL morphs. While the MMPMR for the other three landmark-based methods is higher, those morphs contain very obvious artefacts. Since the MIPGAN optimisation process includes a perception-style loss that encourages visual similarity to both contributing identities, the MIPGAN morphs contain some ghosting artefacts. Because we do not include such a loss during optimisation phase 2, the improved MIPGAN morphs are visually more convincing than MIPGAN morphs and landmark morphs that contain visible ghosting artefacts. Some of the other three landmark-based approaches outperform improved MIPGAN, but contain significant ghosting artefacts that would not fool visual inspection by humans. If a large network such as StyleGAN were explicitly \emph{trained} to generate morphs, they might become even more challenging.

\begin{table*}[t]
	\caption{Detection performance in BPCER (\%) at APCER$\leq$5\% and $\leq$10\% (Section \ref{MAD}). Top: LBP-based S-MAD. Bottom: D-MAD based on FR-difference features.}\label{tabmad}
	\centering
		\begin{tabular}{|l|l|l|l|l|l|l|l|l|l|l|l|}
			\hline
			\vspace{-0.3cm}
			& \multicolumn{11}{c|}{}\\
			\multicolumn{1}{|c|}{\textbf{S-MAD}} & \multicolumn{11}{c|}{BPCER@APCER$\leq$5\%}\\
			\hline
			&		&			&	Improved	& \multisev{c|}{ WALI (Ours) with optimisation using:} & WALI (Ours)\\
			
			& Land-	& MIPGAN	&	MIPGAN		& \multi{c}{ \ \it{MFN} }
			&\multi{c}{\it{MFN}\&} 
			&\multi{c}{\it{MFN}\&}
			&\multi{c}{\it{MFN}\&}
			&\multi{c}{\it{MFN}\&}
			&\multi{c}{\it{MFN}\&}
			&\multi{c|}{\it{MFN}\&} 		& 512$\times$512\\
	Trained with 		& mark	&			&	(Ours)		& \multi{c}{}	&\multi{c}{\it{El.Face}} 
			&\multi{c}{\it{Curr.Face}}
			&\multi{c}{\it{ArcFace}}
			&\multi{c}{\it{Inception}}
			&\multi{c}{\it{PocketNet}} 
			&\multi{c|}{\it{VGG16}} & baseline\\	
			\hline
			FRGC & 3.2	& 100.0	& 99.4	& 71.5	& 66.2	& 67.0	& 69.2	& 70.9	& 71.2	& 63.9	& 81.3 \\
			\hline
			AMSL & 38.7	& 100.0	& 95.2	& 54.2	& 62.4	& 48.8	& 45.9	& 48.4	& 48.0	& 42.6	& 99.8 \\
			\hline
			\vspace{-0.3cm}
			& \multicolumn{11}{c|}{}\\
			& \multicolumn{11}{c|}{BPCER@APCER$\leq$10\%}\\
			\hline
			FRGC & 1.4	& 100.0	& 98.6	& 56.0	& 52.2	& 50.7	& 55.9	& 49.8	& 45.5	& 47.9	& 78.8 \\
			\hline
			AMSL & 26.2	& 100.0	& 89.0	& 29.5	& 38.1	& 31.4	& 26.5	& 26.9	& 26.7	& 26.9	& 99.4 \\
			\hline			
		\end{tabular}
		
		\vspace{0.4cm}
		
		\begin{tabular}{|l|l|l|l|l|l|l|l|l|l|l|l|}
			\hline
			\vspace{-0.3cm}
			& \multicolumn{11}{c|}{}\\
			\multicolumn{1}{|c|}{\textbf{D-MAD}}	& \multicolumn{11}{c|}{BPCER@APCER$\leq$5\%}\\
			\hline
			&		&			&	Improved	& \multisev{c|}{ WALI (Ours) with optimisation using:} & WALI (Ours)\\
			
			& Land-	& MIPGAN	&	MIPGAN		& \multi{c}{ \ \it{MFN} }
			&\multi{c}{\it{MFN}\&} 
			&\multi{c}{\it{MFN}\&}
			&\multi{c}{\it{MFN}\&}
			&\multi{c}{\it{MFN}\&}
			&\multi{c}{\it{MFN}\&}
			&\multi{c|}{\it{MFN}\&} 		& 512$\times$512\\
			Trained with 		& mark	&			&	(Ours)		& \multi{c}{}	&\multi{c}{\it{El.Face}} 
			&\multi{c}{\it{Curr.Face}}
			&\multi{c}{\it{ArcFace}}
			&\multi{c}{\it{Inception}}
			&\multi{c}{\it{PocketNet}} 
			&\multi{c|}{\it{VGG16}}& baseline\\	
			\hline
			FRGC & 0.3	& 19.4	& 18.5	& 7.2	& 9.8	& 10.2	& 12.8	& 8.3	& 6.7	& 8.4	& 0.5	\\
			\hline
			\vspace{-0.3cm}
			& \multicolumn{10}{c|}{}\\
			& \multicolumn{10}{c|}{BPCER@APCER$\leq$10\%}\\
			\hline
			FRGC & 0.2	& 12.4	& 12.0	& 3.5	& 5.1	& 5.6	& 7.3	& 4.8	& 3.7	& 4.2	& 0.2	\\
			\hline			
		\end{tabular}
\end{table*}

\subsection{S-MAD using LBP}
\begin{figure}[t]
	\centering
	\begin{tikzpicture}
	\newcommand\x{1.6}
	\newcommand\w{0.07}
	\newcommand{\shiftleft}{1.0}
	\newcommand{\shiftdown}{1.6}
	\filldraw[black] (-3.0, 0.12) node[anchor=west] {\footnotesize{WALI 128$\times$128}};
	\filldraw[black] (-3.0, -0.12) node[anchor=west] {\footnotesize{finetuned}};
	\node[inner sep=0pt] (whitehead) at (0,0)
	{\includegraphics[width=\w\paperwidth]{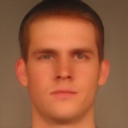}};
	\node[inner sep=0pt] (whitehead) at (\x,0)
	{\includegraphics[width=\w\paperwidth]{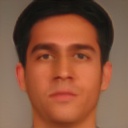}};
	\node[inner sep=0pt] (whitehead) at (\x *2,0)
	{\includegraphics[width=\w\paperwidth]{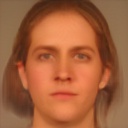}};
	\node[inner sep=0pt] (whitehead) at (\x *3,0)
	{\includegraphics[width=\w\paperwidth]{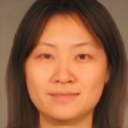}};
	
	\filldraw[black] (-3.0, 0.12 - \shiftdown) node[anchor=west] {\footnotesize{WALI 128$\times$128}};	
	\filldraw[black] (-3.0, -0.12 - \shiftdown) node[anchor=west] {\footnotesize{baseline}};	
	\node[inner sep=0pt] (whitehead) at (0,0 - \shiftdown)
	{\includegraphics[width=\w\paperwidth]{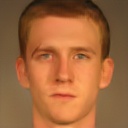}};
	\node[inner sep=0pt] (whitehead) at (\x,0 - \shiftdown)
	{\includegraphics[width=\w\paperwidth]{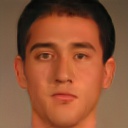}};
	\node[inner sep=0pt] (whitehead) at (\x *2,0 - \shiftdown)
	{\includegraphics[width=\w\paperwidth]{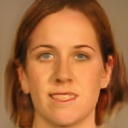}};
	\node[inner sep=0pt] (whitehead) at (\x *3,0 - \shiftdown)
	{\includegraphics[width=\w\paperwidth]{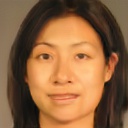}};
	
	\filldraw[black] (-3.0, 0.12 - 2*\shiftdown) node[anchor=west] {\footnotesize{WALI 512$\times$512}};	
	\filldraw[black] (-3.0, -0.12 - 2*\shiftdown) node[anchor=west] {\footnotesize{baseline}};		
	\node[inner sep=0pt] (whitehead) at (0,0 - 2*\shiftdown)
	{\includegraphics[width=\w\paperwidth]{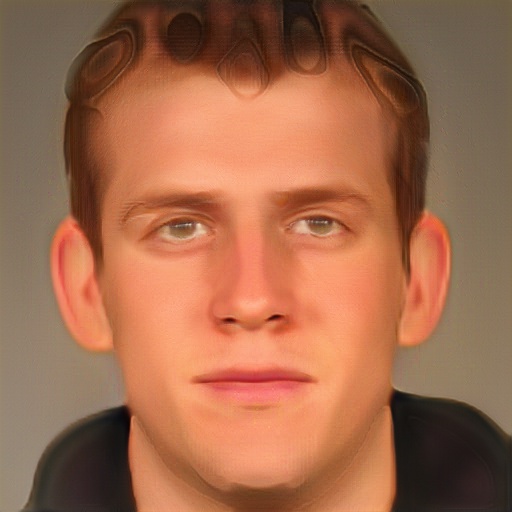}};
	\node[inner sep=0pt] (whitehead) at (\x,0 - 2*\shiftdown)
	{\includegraphics[width=\w\paperwidth]{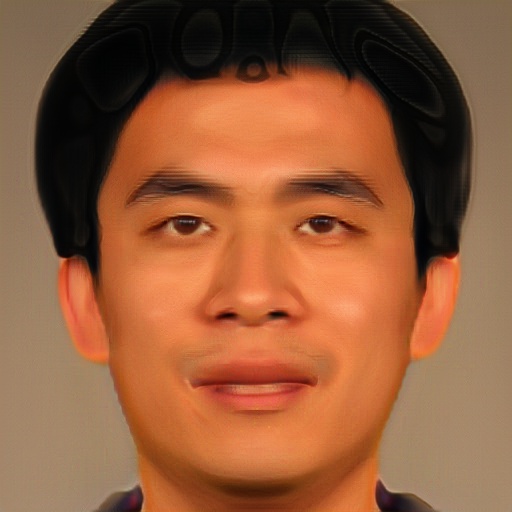}};
	\node[inner sep=0pt] (whitehead) at (\x *2,0 - 2*\shiftdown)
	{\includegraphics[width=\w\paperwidth]{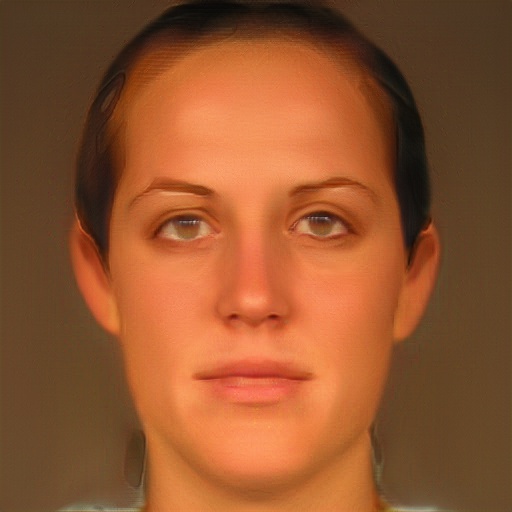}};
	\node[inner sep=0pt] (whitehead) at (\x *3,0 - 2*\shiftdown)
	{\includegraphics[width=\w\paperwidth]{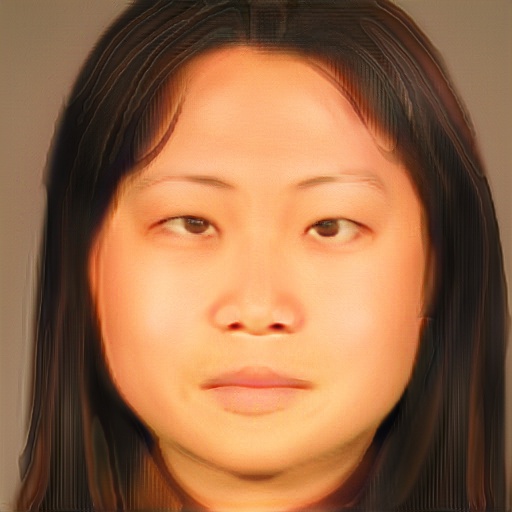}};
	
	\end{tikzpicture}
	\caption{\label{fig:WALI512}Comparison of morphs generated by a 128$\times$128 WALI model trained with FR losses (top row), a 128$\times$128 WALI model trained without FR losses (second row) and a 512$\times$512 WALI model without FR losses (bottom row). Comparing the top two rows shows that there seems to be a trade-off between image quality and morphing performance. Comparing the two bottom rows shows that blurriness can be corrected for by simply generating higher-resolution images.}
\end{figure}
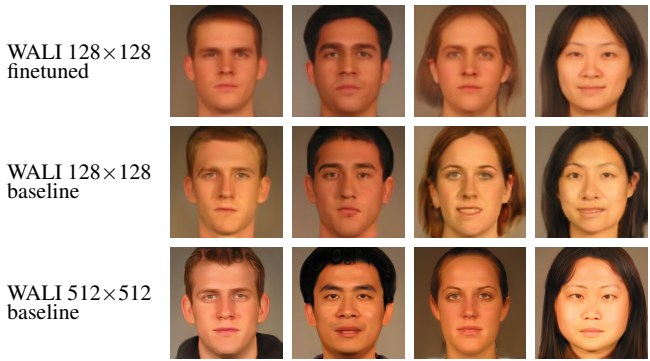

We implement an S-MAD approach based on Local Binary Patterns (LBP) followed by a Support Vector Machine (SVM). We compare the ability of this model to detect morphs: once when it was trained using the same training data as WALI and once when using a separate training set. LBP features may be appropriate for detecting WALI-based morphs when the underlying training data is known, but performance decreases significantly when the database is unknown and contains only landmark-based morphs, see Table \ref{tabmad} and Fig. \ref{fig:DET}. LBP features are not at all suitable for detecting (improved) MIPGAN morphs, which is probably due to the ability of StyleGAN to generate images with texture that is similar to that of real images. The APCER for images generated by a 512$\times$512 WALI model trained without FR losses, see the bottom row in Fig. \ref{fig:WALI512}) and the last column in Table \ref{tabmad}, ranges from 78.8\% to 99.8\%, showing a similar effect.

\subsection{D-MAD using deep-learning-based FR feature differences}
\begin{figure}[h]
	\includegraphics[width=0.45\paperwidth]{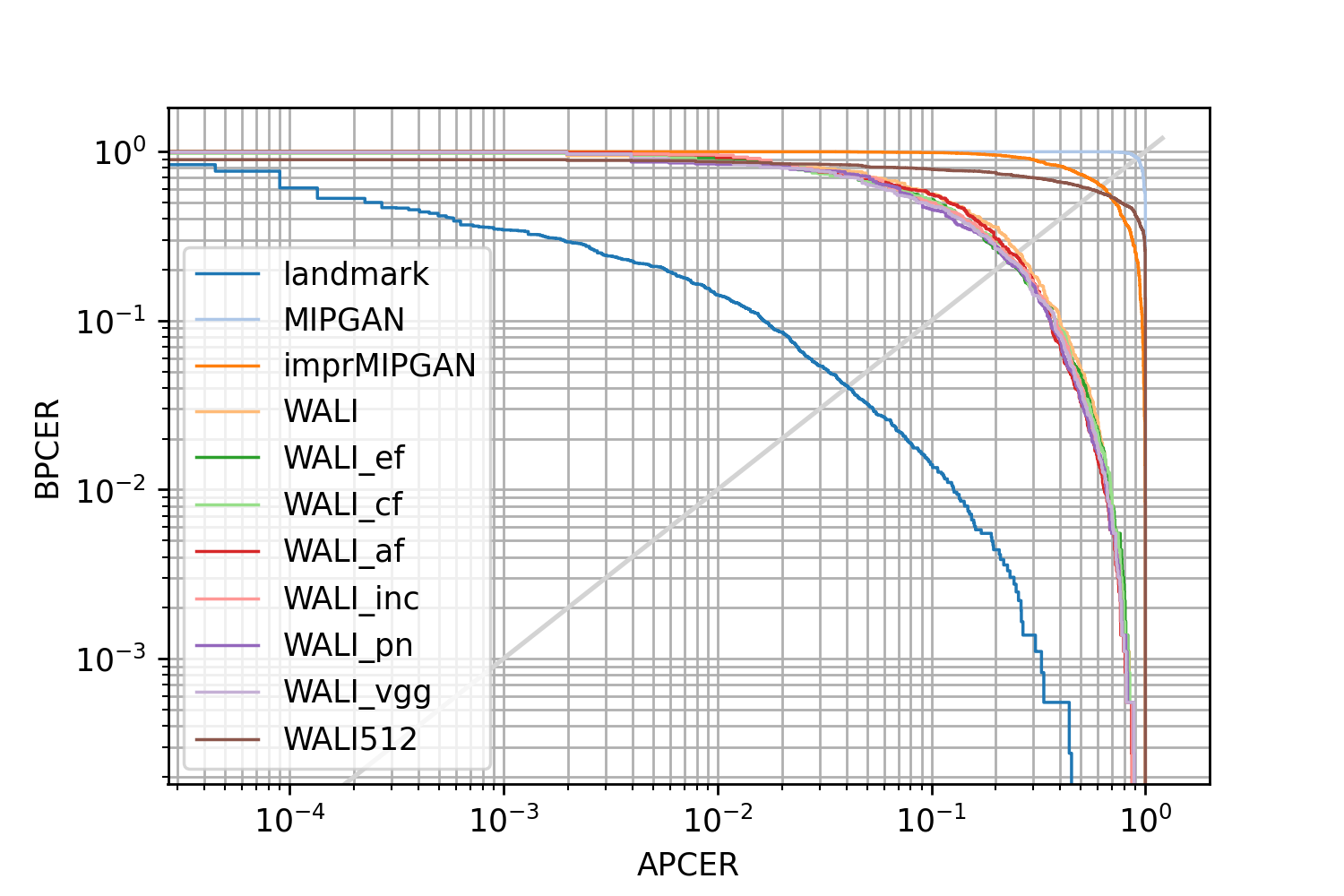}
	\includegraphics[width=0.45\paperwidth]{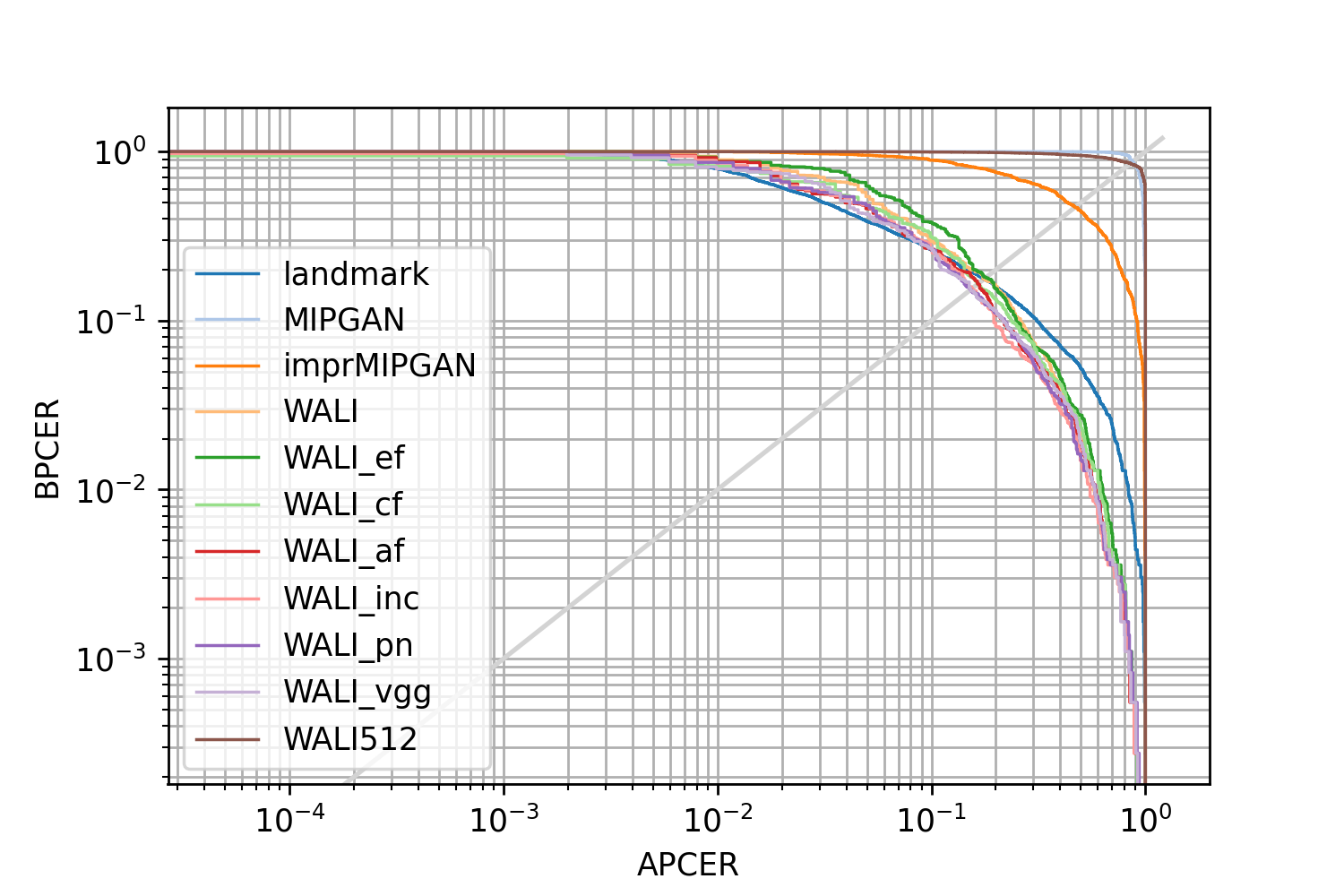}
	\includegraphics[width=0.45\paperwidth]{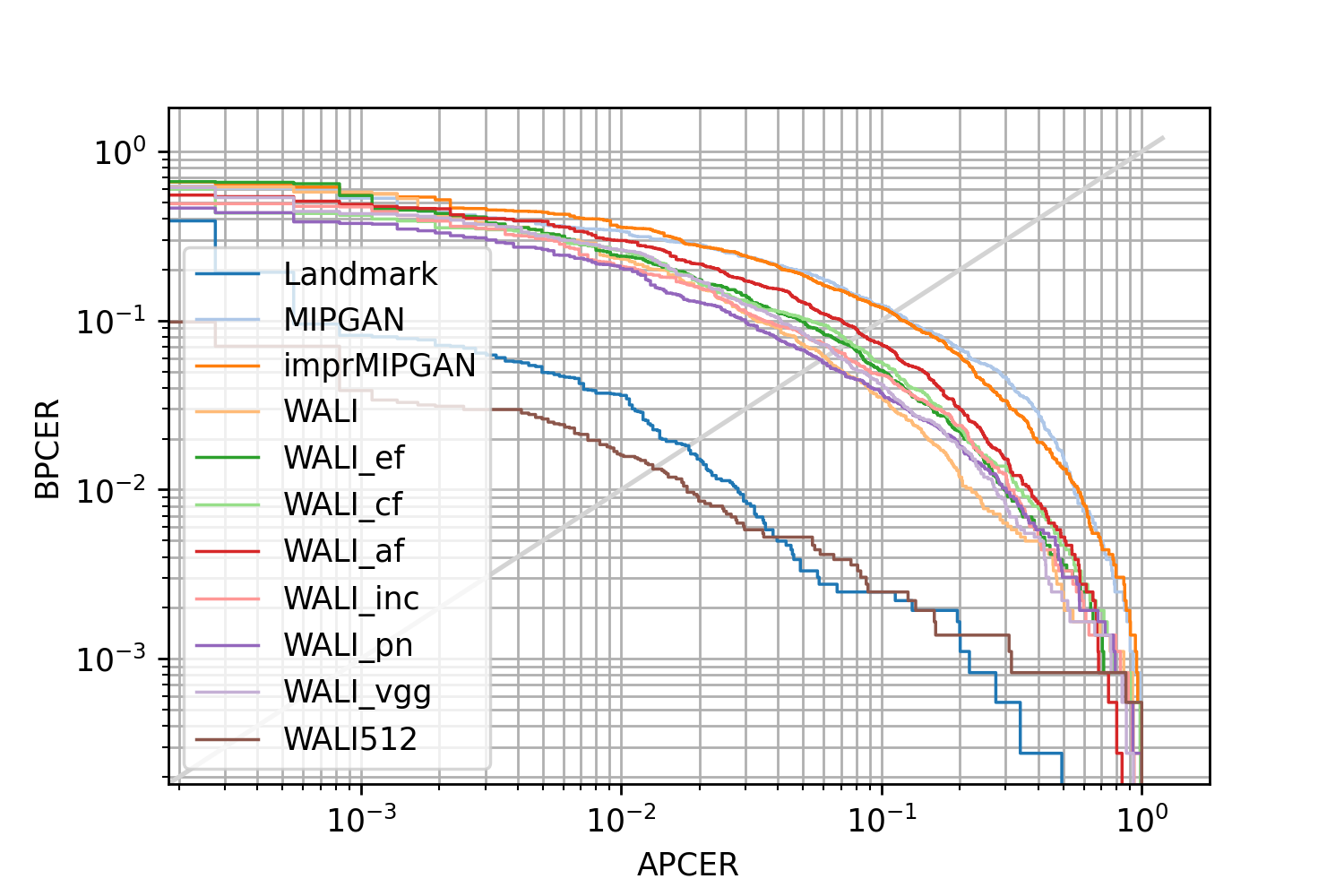}	
	\caption{\label{fig:DET}DET curves. Top: LBP-based SMAD trained with FRGC data. Middle: LBP-based SMAD trained with FRLL data. Bottom: DMAD based on FR feature difference trained with FRGC data.}
\end{figure}

While this approach seems to be very successful at detecting morphed images that were created using the same algorithm that was used to create the training set, its performance decreases significantly when evaluating (improved) MIPGAN or WALI morphs, see Table \ref{tabmad} and Fig. \ref{fig:DET}. Note that this D-MAD approach can detect images generated by a 512$\times$512 WALI model trained without FR losses much more easily than other GAN-based morphs, which makes sense, since these morphs were not optimised using FR systems. If this approach were trained with a separate training set other than FRGC we would expect its performance on MIPGAN or WALI morphs to decrease further.

\subsection{Discussion}
For all FR systems we evaluated, except Dlib, our approach outperforms MIPGAN morphs based on FRGC. For three out of six FR systems tested under black-box assumptions WALI morph outperform landmark morphs. This shows that it is possible to approximate the theoretical worst case for more than one FR system. As we already mentioned, this does not mean that ElasticFace and CurricularFace are generally less vulnerable to GAN-based morphing attacks. These two FR systems are newer and seem to show different behaviour from the other FR systems we tested.

Using WALI to generate morphs is computationally expensive, since optimisation needs to be performed for every morph that is generated. Due to hardware limitations we report results for $128\times128$ images. While we did successfully generate larger images - up to $512\times512$, compared to MIPGAN morphs that rely on a StyleGAN model that generates $1024\times1024$ images - this takes significantly longer and requires more GPU memory, especially during training. However, our results do show that morphs exist that are extremely successful at exploiting the vulnerabilities of (multiple) FR systems. Therefore, the idea of a criminal tweaking their morph in ways to make it more likely to be accepted by multiple FR systems is very possible, illustrating the need to focus on quality as well as quantity when generating morphing datasets. We evaluated five different FR systems and showed that in the (theoretical) worst case up to 72-98\% of FRGC morphs can trick the FR system. For four out of five FR systems we evaluated, our WALI morphs when optimised with two FR systems are closer to this upper bound than either landmark or MIPGAN morphs. As has been reported before \cite{ColboisGAN}, there seems to be an inverse relationship between performance of FR systems on normal data and vulnerability to morphing attacks.

WALI's improvements are due to having a worst-case embedding as a goal to approximate, improved optimisation in two phases (finding a good initial embedding for each bona fide image before generating morphs), optimising with more than one FR system simultaneously, and including the goal of morphing during training. The first three goals can be applied to other existing generative methods, we used StyleGAN as an example, leading to an improved MIPGAN approach that led to morphs that are more challenging than other GAN-based morphs.

WALI morphs were generated in an adversarial manner and probably exploit the fact that deep-learning-based FR systems are sensitive to certain patterns in images. While such patterns might be imperceptible to humans, they can make the FR systems vulnerable to WALI morphs. These patters may not survive post-processing such as printing\&scanning, resizing etc. Furthermore, there are still artefacts visible to the human eye, as can be seen in Fig. \ref{fig:worst-case} and \ref{fig:WALI512}, for example around the mouth or eyes. Visual inspection would probably allow e.g. border guards to detect that the generated morph is not a real image. Our findings therefore show room for improvement for FR systems. We hope that our proposed method WALI can contribute to such an improvement by generating more challenging training data for FR systems.

\section{Conclusion \& Future Work} \label{conclusion}

In this work, we showed that generating challenging morphs is possible and necessary to evaluate the robustness of FR systems. Our newly proposed WALI method outperformed existing morphing techniques on FRGC data, and since it provides a way to generate large quantities of difficult morphs, it could contribute to improving FR and MAD systems' performance. We also introduced an improved MIPGAN approach that due to the powerful underlying StyleGAN Generator generated challenging morphs on FRLL as well as on FRGC. We showed that if the goal of generating challenging morphs is not explicitly considered during the training of a GAN, then the resulting morphs will be significantly less challenging than when that goal \textit{is} included during training. 

Challenges for future research include generating such datasets while also making sure to cover the possible range of morphs by focussing on (visual) quality as well as quantity, for example by investigating the effect of time-consuming manual post-processing. It would be interesting to explore whether GAN networks that can produce images with as high quality as e.g. StyleGAN can also be adapted to explicitly include the goal of generating difficult morphs during training. We showed that optimising towards a worst case leads to more challenging morphs, similar adaptations could be made to Diffusion-based approaches as well. Additionally, further investigation could be carried out on the effect of post-processing techniques on the robustness of FR systems to morphs. Moreover, the effects of training FR systems or MAD methods with large datasets generated with WALI or improved MIPGAN could be further explored in future research.

\section{Ethics, Broader Impact, and Reproducibility}
This paper introduces methods to generate morphs, which could potentially be used to apply for passports or other documents that could be shared by two people, for example allowing them to avoid travel restrictions. As long as countries allow applicants to provide their own digital or printed passport photo, this will continue to pose a risk. On the other hand, sharing our morphing generation method will allow researchers to be more aware of potential vulnerabilities, and support development of countermeasures. Our method can be used to generate large datasets of advanced morphs that can for example be used to train FR systems or to teach human border control staff to better spot morph-related artefacts. We aim to raise awareness for risks posed by morphing and without sharing our method, such vulnerabilities might remain unknown. We also intend to share our code for research purposes only.
To aid reproducibility, we have included important information such as hyperparameters in this paper. All data we used is already available to researchers and we plan to release our code for research purposes after publication.

\section{Acknowledgements}
This work was funded by the Rijksdienst voor Identiteitsgegevens (RvIG). We thank Prof. Dr. Christoph Brune and Dr. Jelmer Wolterink for useful discussions.

\bibliographystyle{ieeetran}
\bibliography{WALI}

\end{document}